\documentclass[journal]{IEEEtran}
\usepackage[noadjust]{cite}

\usepackage{mhchem}

\ifCLASSINFOpdf
 \usepackage[pdftex]{graphicx}
\else
\fi
\ifCLASSOPTIONcompsoc
  \usepackage[caption=false,font=normalsize,labelfont=sf,textfont=sf]{subfig}
\else
  \usepackage[caption=false,font=footnotesize]{subfig}
\fi
\begin{document}
%
\title{A Survey of Neuromorphic Computing and Neural Networks in Hardware}
%
%
%

\author{Catherine D. Schuman,~\IEEEmembership{Member,~IEEE,}
\thanks{C.D. Schuman is with the Computational Data Analytics Group, Oak Ridge National Laboratory, Oak Ridge,
TN, 37831 USA e-mail: schumancd@ornl.gov} Thomas E. Potok,~\IEEEmembership{Member,~IEEE,} Robert M. Patton,~\IEEEmembership{Member,~IEEE,} J. Douglas Birdwell,~\IEEEmembership{Fellow,~IEEE,} Mark E. Dean,~\IEEEmembership{Fellow,~IEEE,} Garrett S. Rose,~\IEEEmembership{Member,~IEEE,} and James S. Plank,~\IEEEmembership{Member,~IEEE}}

\maketitle

\begin{abstract}
Neuromorphic computing has come to refer to a variety of brain-inspired computers, devices, and models that contrast the pervasive von Neumann computer architecture. This biologically inspired approach has created highly connected synthetic neurons and synapses that can be used to model neuroscience theories as well as solve challenging machine learning problems. The promise of the technology is to create a brain-like ability to learn and adapt, but the technical challenges are significant, starting with an accurate neuroscience model of how the brain works, to finding materials and engineering breakthroughs to build devices to support these models, to creating a programming framework so the systems can learn, to creating applications with brain-like capabilities. In this work, we provide a comprehensive survey  of the research and motivations for neuromorphic computing over its history.  We begin with a 35-year review of the motivations and drivers of neuromorphic computing, then look at the major research areas of the field, which we define as neuro-inspired models, algorithms and learning approaches,  hardware and devices, supporting systems, and finally applications. We conclude with a broad discussion on the major research topics that need to be addressed in the coming years to see the promise of neuromorphic computing fulfilled. The goals of this work are to provide an exhaustive review of the research conducted in neuromorphic computing since the inception of the term, and to motivate further work by illuminating gaps in the field where new research is needed.\let\thefootnote\relax\footnote{Notice: This manuscript has been authored by UT-Battelle, LLC under Contract No. DE-AC05-00OR22725 with the U.S. Department of Energy. The United States Government retains and the publisher, by accepting the article for publication, acknowledges that the United States Government retains a non-exclusive, paid-up, irrevocable, world-wide license to publish or reproduce the published form of this manuscript, or allow others to do so, for United States Government purposes. The Department of Energy will provide public access to these results of federally sponsored research in accordance with the DOE Public Access Plan (http://energy.gov/downloads/doe-public-access-plan).} 
\end{abstract}


%
\IEEEpeerreviewmaketitle

\section{Introduction}
%
%
%
%
\IEEEPARstart{T}{his} paper provides a comprehensive survey of the neuromorphic computing field, reviewing over 3,000 papers from a 35-year time span looking primarily at the motivations, neuron/synapse models, algorithms and learning, applications, advancements in hardware, and briefly touching on materials and supporting systems. Our goal is to provide a broad and historic perspective of the field to help further ongoing research, as well as provide a starting point for those new to the field.

Devising a machine that can process information faster than humans has been a driving forces in computing for decades, and the von Neumann architecture has become the clear standard for such a machine. However, the inevitable comparisons of this architecture to the human brain highlight significant differences in the organizational structure, power requirements, and processing capabilities between the two. This leads to a natural question regarding the feasibility of creating alternative architectures based on neurological models, that compare favorably to a biological brain. 

\begin{figure}[!t]
\begin{center}
\includegraphics[width=0.5\textwidth]{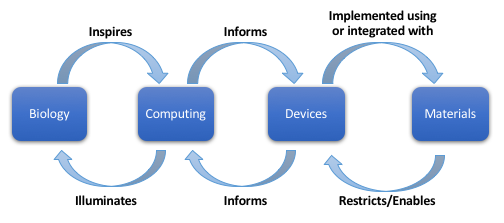}
\caption{Areas of research involved in neuromorphic computing, and how they are related.}
\label{fig:areas_of_research}
\end{center}
\end{figure}

Neuromorphic computing has emerged in recent years as a complementary architecture to von Neumann systems. The term neuromorphic computing was coined in 1990 by Carver Mead \cite{mead1990}.  At the time, Mead referred to very large scale integration (VLSI) with analog components that mimicked biological neural systems as ``neuromorphic" systems.  More recently, the term has come to encompass implementations that are based on biologically-inspired or artificial neural networks in or using non-von Neumann architectures.  

These neuromorphic architectures are notable for being highly connected and parallel, requiring low-power, and collocating memory and processing. While interesting in their own right, neuromorphic architectures have received increased attention due to the approaching end of Moore's law, the increasing power demands associated with Dennard scaling, and the low bandwidth between CPU and memory known as the von Neumann bottleneck \cite{monroe2014neuromorphic}. Neuromorphic computers have the potential to perform complex calculations faster, more power-efficiently, and on a smaller footprint than traditional von Neumann architectures.  These characteristics provide compelling reasons for developing hardware that employs neuromorphic architectures.

Machine learning provides the second important reason for strong interest in neuromorphic computing. The approach shows promise in improving the overall learning performance for certain tasks. This moves away from hardware benefits to understanding potential application benefits of neuromorphic computing, with the promise of developing algorithms that are capable of on-line, real-time learning similar to what is done in biological brains. Neuromorphic architectures appear to be the most appropriate platform for implementing machine learning algorithms in the future.

The neuromorphic computing community is quite broad, including researchers from a variety of fields, such as materials science, neuroscience, electrical engineering, computer engineering, and computer science (Figure \ref{fig:areas_of_research}).  Materials scientists study, fabricate, and characterize new materials to use in neuromorphic devices, with a focus on materials that exhibit properties similar to biological neural systems.  Neuroscientists provide information about new results from their studies that may be useful in a computational sense, and utilize neuromorphic systems to simulate and study biological neural systems.  Electrical and computer engineers work at the device level with analog, digital, mixed analog/digital, and non-traditional circuitry to build new types of devices, while also determining new systems and communication schemes. Computer scientists and computer engineers work to develop new network models inspired by both biology and machine learning, including new algorithms that can allow these models to be trained and/or learn on their own.  They also develop the supporting software necessary to enable the use of neuromorphic computing systems in the real world. 

The goals of this paper are to give a thirty-year survey of the published works in neuromorphic computing and hardware implementations of neural networks and to discuss open issues for the future of neuromorphic computing.  The remainder of the paper is organized as follows:  In Section \ref{sec:motivation}, we present a historical view of the  motivations for developing neuromorphic computing and how they have changed over time.  We then break down the discussion of past works in neuromorphic computing into models (Section \ref{sec:models}), algorithms and learning (Section \ref{sec:algorithms}),  hardware implementations (Section \ref{sec:hardware}), and supporting components, including communication schemes and software systems (Section \ref{sec:supporting}).  Section \ref{sec:applications} gives an overview of the types of applications to which neuromorphic computing systems have been applied in the literature.  Finally, we conclude with a forward-looking perspective for neuromorphic computing and enumerate some of the major research challenges that are left to tackle.

\begin{figure}[!t]
\begin{center}
\includegraphics[width=0.5\textwidth]{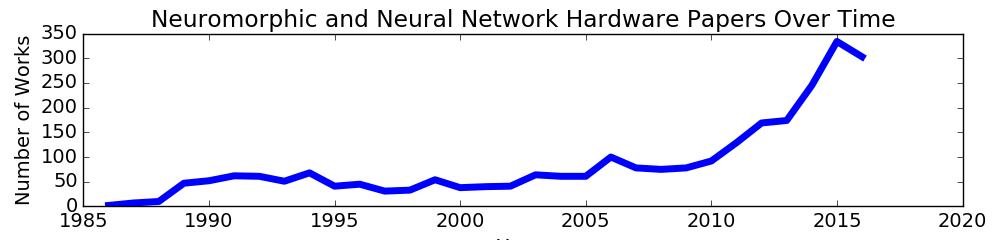}
\caption{Neuromorphic and neural network hardware works over time.}
\label{fig:works_over_time}
\end{center}
\end{figure}

\section{Motivation}
\label{sec:motivation}

\begin{figure*}[!t]
\centering
\includegraphics[width=0.9\textwidth]{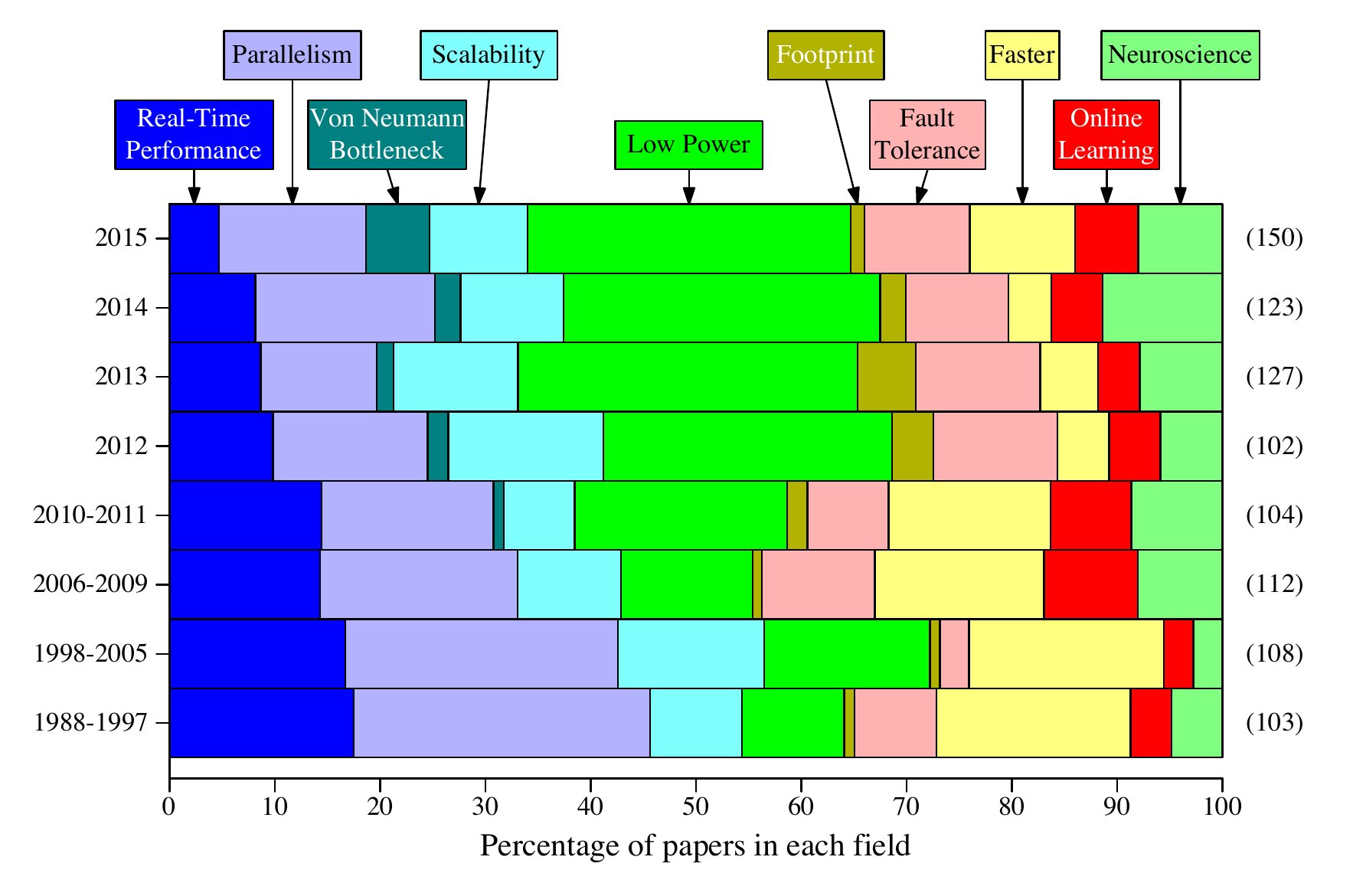}
\caption{Ten different motivations for developing neuromorphic systems, and over time, the percentage of the papers in the literature that have indicated that motivation as one of the primary reasons they have pursued the development of neuromorphic systems.}
\label{fig:motivations_over_time}
\end{figure*}

The idea of using custom hardware to implement neurally-inspired systems is as old as computer science and computer engineering itself, with both von Neumann \cite{von2012the} and Turing \cite{turing1950computing} discussing brain-inspired machines in the 1950's.  Computer scientists have long wanted to replicate biological neural systems in computers.  This pursuit has led to key discoveries in the fields of artificial neural networks (ANNs), artificial intelligence, and machine learning.  The focus of this work, however, is not directly on ANNs or neuroscience itself, but on the development of non-von Neumann hardware for simulating ANNs or biological neural systems.   We discuss several reasons why neuromorphic systems have been developed over the years based on motivations described in the literature.  Figure \ref{fig:works_over_time} shows the number of works over time for neuromorphic computing and indicates that there has been a distinct rise in interest in the field over the last decade. Figure \ref{fig:motivations_over_time} shows ten of the primary motivations for neuromorphic in the literature and how those motivations have changed over time. These ten motivations were chosen because they were the most frequently noted motivations in the literature, each specified by at least fifteen separate works. 

Much of the early work in neuromorphic computing was spurred by the development of hardware that could perform parallel operations, inspired by observed parallelism in biological brains, but on a single chip \cite{murray1988, blayo1989a, salam1989a, bibyk1990, distante1990b}.  
Although there were parallel architectures available,  neuromorphic systems emphasized many simple processing components (usually in the form of neurons), with relatively dense interconnections between them (usually in the form of synapses), differentiating them from other parallel computing platforms of that time.  In works from this early era of neuromorphic computing, the inherent parallelism of neuromorphic systems was the most popular reason for custom hardware implementations. 

Another popular reason for early neuromorphic and neural network hardware implementations was speed of computation \cite{burr1991, chiang1991, madani1991, murray1991}.  In particular, developers of early systems emphasized that it was possible to achieve much faster neural network computation with custom chips than what was possible with traditional von Neumann architectures, partially by exploiting their natural parallelism as mentioned above, but also by building custom hardware to complete neural-style computations.  This early focus on speed foreshadowed a future of utilizing neuromorphic systems as accelerators for machine learning or neural network style tasks.  

Real-time performance was also a key motivator of early neuromorphic systems.  Enabled by natural parallelism and speed of computation, these devices tended to be able to complete neural network computations faster than implementations on von Neumann architectures for applications such as real-time control \cite{hasler1990}, real-time digital image reconstruction \cite{lee1990b}, and autonomous robot control \cite{tarassenko1991}.  In these cases, the need for faster computation was not driven by studying the neural network architectures themselves or for training, but was more driven by application performance.  This is why we have differentiated it from speed and parallelism as a motivation for the development of neuromorphic systems.  

Early developers also started to recognize that neural networks may be a natural model for hardware implementation because of their inherent fault tolerance, both in the massively parallel representation and in potential adaptation or self-healing capabilities that can be present in artificial neural network representations in software \cite{murray1988, akers1989}.  These were and continue to be relevant characteristics for fabricating new hardware implementations, where device and process variation may lead to imperfect fabricated devices, and where utilized devices may experience hardware errors. 


The most popular motivation in present-day literature and discussions of neuromorphic systems in the referenced articles is the emphasis on their potential for extremely low power operation.  Our major source of inspiration, the human brain, requires about 20 watts of power and performs extremely complex computations and tasks on that small power budget.  The desire to create neuromorphic systems that consume similarly low power has been a driving force for neuromorphic computing from its conception \cite{leong1992, cairns1994}, but it became a prominent motivation about a decade into the field's history. 

Similarly, creating devices capable of neural network-style computations with a small footprint (in terms of device size) also became a major motivation in this decade of neuromorphic research.  Both of these motivations correspond with the rise of the use of embedded systems and microprocessors, which may require a small footprint and, depending on the application, very low power consumption.  

In recent years, the primary motivation for the development of neuromorphic systems is low-power consumption.  It is, by far, the most popular motivation for neuromorphic computers, as can be seen in Figure \ref{fig:motivations_over_time}.  Inherent parallelism, real-time-performance, speed in both operation and training, and small device footprint also continue to be major motivations for the development of neuromorphic implementations.  A few other motivations became popular in this period, including a rise of approaches that utilize neural network-style architectures (i.e., architectures made up of neuron and synapse-like components) because of their fault tolerance characteristics or reliability in the face of hardware errors.  This has become an increasingly popular motivation in recent years in light of the use of novel materials for implementing neuromorphic systems (see Section \ref{sec:materials}).  

Another major motivation for building neuromorphic systems in recent years has been to study neuroscience.  Custom neuromorphic systems have been developed for several neuroscience-driven projects, including those created as part of the European Union's Human Brain Project \cite{markram2012human}, because simulating relatively realistic neural behavior on a traditional supercomputer is  not feasible, in scale, speed, or power consumption \cite{schemmel2010}. 
As such, custom neuromorphic implementations are required in order to perform meaningful neuroscience simulations with reasonable effort.  In this same vein, scalability has also become an increasingly popular motivation for building neuromorphic systems.  Most major neuromorphic projects discuss how to cascade their devices together to reach very large numbers of neurons and synapses.  

A common motivation not given explicitly in Figure \ref{fig:motivations_over_time} is the end of Moore's Law, though most of the other listed motivations are related to the consideration of neuromorphic systems as a potential complementary architecture in the beyond Moore's law computing landscape.  Though most researchers do not expect that neuromorphic systems will replace von Neumann architectures, ``building a better computer" is one of their motivations for developing neuromorphic devices; though this is a fairly broad motivation, it encompasses issues associated with traditional computers, including the looming end of Moore's law and the end of Dennard scaling.  Another common motivation for neuromorphic computing development is solving the von Neumann bottleneck \cite{backus1978can}, which arises in von Neumann architectures due to the separation of memory and processing and the gap in performance between processing and memory technologies in current systems.  In neuromorphic systems, memory and processing are collocated, mitigating issues that arise with the von Neumann bottleneck. 

On-line learning, defined as the ability to adapt to changes in a task as they occur, has been a key motivation for neuromorphic systems in recent years.  Though on-line learning mechanisms are still not well understood, there is still potential for the on-line learning mechanisms that are present in many neuromorphic systems to perform learning tasks in an unsupervised, low-power manner.  With the tremendous rise of data collection in recent years, systems that are capable of processing and analyzing this data in an unsupervised, on-line way will be integral in future computing platforms.  Moreover, as we continue to gain an understanding of biological brains, it is likely that we will be able to build better on-line learning mechanisms and that neuromorphic computing systems will be natural platforms on which to implement those systems.


\section{Models}
\label{sec:models}
One of the key questions associated with neuromorphic computing is which neural network model to use.  The neural network model defines what components make up the network, how those components operate, and how those components interact.  For example, common components of a neural network model are neurons and synapses, taking inspiration from biological neural networks. When defining the neural network model, one must also define models for each of the components (e.g., neuron models and synapse models); the component model governs how that component operates.  

How is the correct model chosen? In some cases, it may be that the chosen model is motivated by a particular application area.   For example, if the goal of the neuromorphic device is to utilize the device to simulate biological brains for a neuroscience study on a faster scale than is possible with traditional von Neumann architectures, then a biologically realistic and/or plausible model is necessary.  If the application is an image recognition task that requires high accuracy, then a neuromorphic system that implements convolutional neural networks may be best.  The model itself may also be shaped by the characteristics and/or restrictions of a particular device or material.  For example, memristor-based systems (discussed further in Section \ref{sec:memristors}) have characteristics that allow for spike-timing dependent plasticity-like mechanisms (a type of learning mechanism discussed further in Section \ref{sec:algorithms}) that are most appropriate for spiking neural network models.  In many other cases, the choice of the model or the level of complexity for the model is not entirely clear.  

A wide variety of model types have been implemented in neuromorphic or neural network hardware systems. The models range from predominantly biologically-inspired to predominantly computationally driven.  The latter models are inspired more by artificial neural network models than by biological brains.  This section discusses different neuron models, synapse models, and network models that have been utilized in neuromorphic systems, and points to key portions of the literature for each type of model. 

\subsection{Neuron Models}

A biological neuron is usually composed of a cell body, an axon, and dendrites.  The axon usually (though not always) transmits information away from the neuron, and is where neurons transmit output.  Dendrites usually (though not always) transmit information to the cell body and are typically where neurons receive input.  Neurons can receive information through chemical or electrical transmissions from other neurons.  The juncture between the end of an axon of one neuron and the dendrite of another neuron that allows information or signals to be transmitted between the two neurons is called a synapse. The typical behavior of a neuron is to accumulate charge through a change in voltage potential across the neuron's cell membrane, caused by receiving signals from other neurons through synapses.  The voltage potential in a neuron may reach a particular threshold, which will cause the neuron to ``fire" or, in the biological terminology, generate an action potential that will travel along a neuron's axon to affect the charge on other neurons through synapses.  Most neuron models implemented in neuromorphic systems have some concept of accumulation of charge and firing to affect other neurons, but the mechanisms by which these processes take place can vary significantly from model to model.  Similarly, models that are not biologically plausible (i.e.~artificial models that are \textit{inspired} by neuroscience rather than \textit{mimicking} neuroscience) typically do not implement axons or dendrites, although there are a few exceptions (as noted below).

\begin{figure}[!t]
\centering
\includegraphics[width=0.5\textwidth]{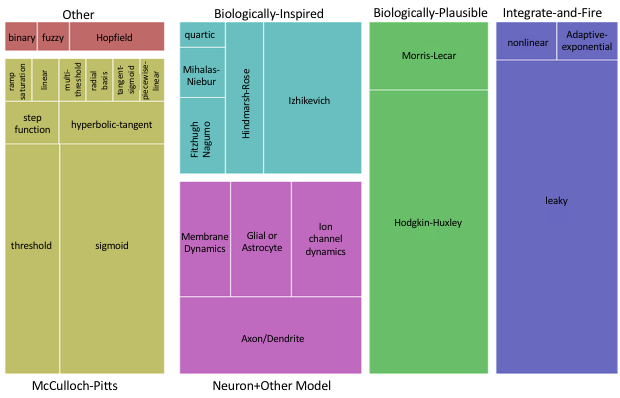}
\caption{A hierarchy of neuron models that have hardware implementations.  The size of the boxes corresponds to the number of implementations for that model, and the color of the boxes corresponds to the ``family" of neuron models, which are labeled either above or below the group of same-colored boxes.}
\label{fig:neuron_model_hierarchy}
\end{figure}

Figure \ref{fig:neuron_model_hierarchy} gives an overview of the types of neuron models that have been implemented in hardware.  The neuron models are given in five broad categories:
\begin{itemize}
\item Biologically-plausible: Explicitly model the types of behavior that are seen in biological neural systems.
\item Biologically-inspired: Attempt to replicate behavior of biological neural systems but not necessarily in a biologically-plausible way.
\item  Neuron+Other: Neuron models including other biologically-inspired components that are not usually included in other neuromorphic neuron models, such as axons, dendrites, or glial cells.
\item Integrate-and-fire: A simpler category of biologically-inspired spiking neuron models.
\item McCulloch-Pitts: Neuron models that are derivatives of the original McCulloch-Pitts neuron \cite{mcculloch1943logical} used in most artificial neural network literature. For this model, the output of neuron $j$ is governed by the following equation:
\begin{equation}
y_j = f \left(\sum_{i=0}^N w_{i,j}x_i\right),
\end{equation}
where $y_j$ is the output value, $f$ is an activation function, $N$ is the number of inputs into neuron $j$, $w_{i,j}$ is the weight of the synapse from neuron $i$ to neuron $j$ and $x_i$ is the output value of neuron $i$. 
\end{itemize}

A variety of biologically-plausible and biologically-inspired neuron models have been implemented in hardware.   Components that may be included in these models may include: cell membrane dynamics, which govern factors such as charge leakage across the neuron's cell membrane; ion channel dynamics, which govern how ions flow into and out of a neuron, changing the charge level of the neuron; axonal models, which may include delay components; and dendritic models, which govern how many pre- and post-synaptic neurons affect the current neuron.  A good overview of these types of spiking neuron models is given by Izhikevich \cite{izhikevich2004}. 

\subsubsection{Biologically-Plausible}

The most popular biologically-plausible neuron model is the Hodgkin-Huxley model \cite{hodgkin1952quantitative}.  The Hodgkin-Huxley model was first introduced in 1952 and is a relatively complex neuron model, with four-dimensional nonlinear differential equations describing the behavior of the neuron in terms of the transfer of ions into and out of the neuron. Because of their biological plausibility, Hodgkin-Huxley models have been very popular in neuromorphic implementations that are trying to accurately model biological neural systems \cite{basu2008, basu2012, castanos2015, castanos2016, deweerth2007, dupeyron1996, grassia2012, grattarola1995, hegab2015, hynna2007, kanoh2005, kohno2008, kohno2008a, lam2015, le1999, ma2012, ma2012a, mahowald1991, parodi1997, pelayo1997, rasche1998, saighi2008, sekikawa2008, shin1999, simoni1998, simoni2001, simoni2004, tomas2010, toumazou1998}.  A simpler, but still biologically-plausible model is the Morris Lecar model, which reduces the model to a two-dimensional nonlinear equation \cite{borisyuk2015morris}.  It is a commonly implemented model in neuroscience and in neuromorphic systems \cite{basu2012, gholami2015, hayati2015a, nakada2012, patel1997, patel2000, sekerli2004}.

\subsubsection{Biologically-Inspired}

There are a variety of neuron models that are simplified versions of the Hodgkin-Huxley model that have been implemented in hardware, including Fitzhugh-Nagumo \cite{binczak2006, cosp2008, linares1991} and Hindmarsh-Rose \cite{hayati2016, lee2004, lu2011, lu2013, lu2014} models.  These models tend to be both simpler computationally and simpler in terms of number of parameters, but they become more biologically-inspired than biologically-plausible because they attempt to model behavior rather than trying to emulate physical activity in biological systems.  From the perspective of neuromorphic computing hardware, simpler computation can lead to simpler implementations that are more efficient and can be realized with a smaller footprint.  From the algorithms and learning method perspective, a smaller number of parameters can be easier to set and/or train than models with a large number of parameters.  

The Izhikevich spiking neuron model was developed to produce similar bursting and spiking behaviors as can be elicited from the Hodgkin-Huxley model, but do so with much simpler computation \cite{izhikevich2003}.  The Izhikevich model has been very popular in the neuromorphic literature because of its simultaneous simplicity and ability to reproduce biologically accurate behavior \cite{basu2012, dutra2013, mizoguchi2011, ning2015, ozoguz2011, radhika2015, rangan2010, sharifipoor2012, soleimani2014, wijekoon2006, wijekoon2008, wijekoon2008a, wijekoon2009}. The Mihala\c{s}-Niebur neuron is another popular neuron model that tries to replicate bursting and spiking behaviors, but it does so with a set of linear differential equations \cite{mihalacs2009}; it also has neuromorphic implementations \cite{folowosele2009a, folowosele2011}.  The quartic model has two non-linear differential equations that describe its behavior, and also has an implementation for neuromorphic systems \cite{grassia2014}. 

\subsubsection{Neuron + Other Biologically-Inspired Mechanism}
Other biologically-inspired models are also prevalent that do not fall into the above categories.  They typically contain a much higher level of biological detail than most models from the machine learning and artificial intelligence literature, such as the inclusion of membrane dynamics \cite{arthur2006a, arthur2011, wang2014d, wang2015o, wittig2006}, modeling ion-channel dynamics \cite{basu2010, basu2010b, hynna2001, hynna2006, hynna2007a, meador1989, rasche2000}, the incorporation of axons and/or dendrite models \cite{elias1992, elias1992a, elias1994a, gorelik2003, hasler2007, hussain2016, koch1988, minch1995, rasche2001a, rasche2001b, wang2011b, wang2013a}, and glial cell or astrocyte interactions \cite{hayati2015, irizarry2013, irizarry2015, ranjbar2015, ranjbar2015a, soleimani2015}.  Occasionally, new models are developed specifically with the hardware in mind.  For example, a neuron model with equations inspired by the Fitzhugh-Nagumo, Morris Lecar, Hodgkin-Huxley, or other models have been developed, but the equations were updated or the models abstracted in order to allow for ease of implementation in low-power VLSI \cite{erdener2015, erdener2016}, on FPGA \cite{upegui2003, upegui2004}, or using static CMOS \cite{kohno2014, kohno2014a, kohno2017}. Similarly, other researchers have updated the Hodgkin-Huxley model to account for new hardware developments, such as the MOSFET transistor \cite{farquhar2005, kohno2010, massobrio2007, saeki2002, saeki2006, tete2011, wijekoon2007} or the single-electron transistor \cite{wen2011}.  




%

\subsubsection{Integrate-and-Fire Neurons}

A simpler set of spiking neuron models belong to the integrate-and-fire family, which is a set of models that vary in complexity from relatively simple (the basic integrate-and-fire) to those approaching complexity levels near that of the Izhikevich model and other more complex biologically-inspired models \cite{gerstner2002}.  In general, integrate-and-fire models are less biologically realistic, but produce enough complexity in behavior to be useful in spiking neural systems. The simplest integrate-and-fire model maintains the current charge level of the neuron. There is also a leaky integrate-and-fire implementation that expands the simplest implementation by including a leak term to the model, which causes the potential on a neuron to decay over time.  It is one of the most popular models used in neuromorphic systems \cite{aamir2016, asai2003, bindal2007, bindal2007a, bragg2002, chen2007a, chen2012, chen2016, crebbin2005, deng2015a, folowosele2009, hamilton2011a, iguchi2010, indiveri2003, indiveri2010, lecerf2014, kornijcuk2016, kravtsov2011, li2009, lim2015, liu2001a, liu2003a, livi2009, mattia1997, morie2015, nakada2012, rubin2004, russell2011, russell2012,  srivastava2016, torres2003, wang2012b, wolpert1996}.  The next level of complexity is the general nonlinear integrate-and-fire method, including the quadratic integrate-and-fire model that is used in some neuromorphic systems \cite{basham2009, basham2012}.  Another level of complexity is added with the adaptive exponential integrate-and-fire model, which is similar in complexity to the models discussed above (such as the Izhikevich model).  These have also been used in neuromorphic systems \cite{abbas2014, millner2010}.   

In addition to the previous analog-style spiking neuron models, there are also implementations of digital spiking neuron models.  The dynamics in a digital spiking neuron model are usually governed by a cellular automaton, as opposed to a set of nonlinear or linear differential equations.  A hybrid analog/digital implementation has been created for neuromorphic implementations \cite{hashimoto2010}, as well as implementations of resonate-and-fire \cite{hishiki2009} and rotate-and-fire \cite{hishiki2010, hishiki2011} digital spiking neurons.  A generalized asynchronous digital spiking model has been created in order to allow for exhibition of nonlinear response characteristics \cite{matsubara2011a, matsubara2011b}.  Digital spiking neurons have also been utilized in pulse-coupled networks \cite{torikai2006, torikai2006a, torikai2007, torikai2008, torikai2008a}. Finally, a neuron for a random neural network has been implemented in hardware \cite{cerkez1997}, 

In the following sections, the term spiking neural network will be used to describe full network models.  These spiking networks may utilize any of the above neuron models in their implementation; we do not specify which neuron model is being utilized.  Moreover, in some hardware implementations, such as SpiNNaker (see Section \ref{sec:digital}), the neuron model is programmable, so different neuron models may be realized in a single neuromorphic implementation. 


\subsubsection{McCulloch-Pitts Neurons}

Moving to more traditional artificial neural network implementations in hardware, there is a large variety of implementations of the traditional McCulloch-Pitts neuron model \cite{mcculloch1943logical}.  The perceptron is one implementation of the McCulloch-Pitts model, which uses a simple thresholding function as the activation function; because of its simplicity, it is commonly used in hardware implementations \cite{aunet2004, aunet2008, bohossian1998, draghici1999, taheri1991, varshavsky1998, varshavsky1999, varshavsky1999a, varshavsky1999b, varshavsky2009, zamanlooy2012}.  There has also been significant focus to create implementations of various activation functions for McCulloch-Pitts-style neurons in hardware.  Different activation functions have had varying levels of success in neural network literature, and some activation functions can be computationally intensive.  This complexity in computation can lead to complexity in hardware, resulting in a variety of different activation functions and implementations that are attempting to trade-off complexity and overall accuracy and computational usefulness of the model.  The most popular implementations are the basic sigmoid function \cite{al2008, azizian2011, banuelos2003, acconcia2014, al1998, basaglia1995, chen2006b, del2013, jeyanthi2014, khodabandehloo2012, khodja2010, larkin2006, mishra2007, myers1989a, ortega2014, panicker2012, sahin2012, saichand2008, szabo2004, tsai2015, wilamowski2000} and the hyperbolic tangent function \cite{baptista2013a, lin2008a, santos2011, zamanlooy2014}, but other hardware-based activation functions that have been implemented include the ramp-saturation function \cite{banuelos2003}, linear \cite{jeyanthi2014}, piecewise linear \cite{hikawa2003}, step function \cite{banuelos2003, faridi2017}, multi-threshold \cite{zhu2005}, radial basis function \cite{sahin2012}, the tangent sigmoid function \cite{sahin2012}, and periodic activation functions \cite{merkel2013}.  Because the derivative of the activation function is utilized in the back-propagation training algorithm \cite{rumelhart1988learning}, some circuits implement both the function itself and its derivative, for both sigmoid \cite{ lu2000, lu2000a, armato2011, basterretxea2004, beiu1994a, murtagh1992} and hyperbolic tangent \cite{armato2011}.  A few implementations have focused on creating neurons with programmable activation functions \cite{al1997} or on creating building blocks to construct neurons \cite{lee1995}.  

\begin{figure}[!t]
\centering
\includegraphics[width=0.5\textwidth]{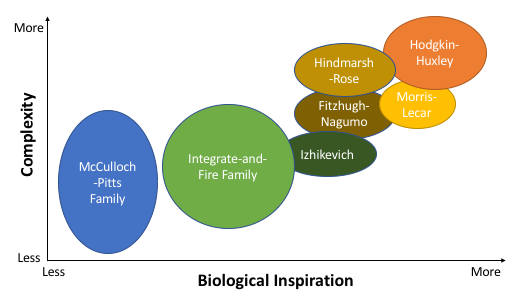}
\caption{A qualitative comparison of neuron models in terms of biological inspiration and complexity of the neuron model.}
\label{fig:neuron_model_inspiration_vs_complexity}
\end{figure}

Neuron models for other traditional artificial neural network models have also been implemented in hardware.  These neuron models include binary neural network neurons \cite{deshmukh2005}, fuzzy neural network neurons \cite{yamakawa1996}, and Hopfield neural network neurons \cite{abo1992, hollis1993, liu2005a}.  On the whole, there have been a wide variety of neuron models implemented in hardware, and one of the decisions a user might make is a tradeoff between complexity and biological inspiration.  Figure \ref{fig:neuron_model_inspiration_vs_complexity} gives a qualitative comparison of different neuron models in terms of those two factors.  

\subsection{Synapse Models}

Just as some neuromorphic work has focused particularly on neuron models, which occasionally also encapsulate the synapse implementation, there has also been a focus on developing synapse implementations independent of neuron models for neuromorphic systems.  Once again, we may separate the synapse models into two categories: biologically-inspired synapse implementations, which include synapses for spike-based systems, and synapse implementations for traditional artificial neural networks, such as feed-forward neural networks.  It is worth noting that synapses are typically going to be the most abundant element in neuromorphic systems, or the element that is going to require the most real estate on a particular chip. For many hardware implementations and especially for the development and use of novel materials for neuromorphic, the focus is typically on optimizing the synapse implementation.  As such, synapse models tend to be relatively simple, unless they are attempting to explicitly model biological behavior.  One popular inclusion for more complex synapse models is a plasticity mechanism, which causes the neuron's strength or weight value to change over time.  Plasticity mechanisms have been found to be related to learning in biological brains. 


For more biologically-inspired neuromorphic networks, synapse implementations that explicitly model the chemical interactions of synapses, such as the ion pumps or neurotransmitter interactions, have been utilized in some neuromorphic systems \cite{friesz2007, gordon2004, gordon2006, kazemi2016, you2016,  lazaridis2006,  thanapitak2013,  lu2011, noack2011, pradyumna2015, pradyumna2015a, wijekoon2011}.  Ion channels have also been implemented in neuromorphic implementations in the form of conductance-based synapse models \cite{bartolozzi2007, benjamin2012, noack2014, rasche1999, shi2003, yu2011b}.  For these implementations, the detail goes above and beyond what one might see with the modeling of ion channels in neuron models such as Hodgkin-Huxley.  

Implementations for spiking neuromorphic systems focus on a variety of characteristics of synapses.  Neuromorphic synapses that exhibit plasticity and learning mechanisms inspired by both short-term and long-term potentiation and depression in biological synapses have been common in biologically-inspired neuromorphic implementations \cite{choi2011, chou2015, covi2015, desbief2015, gopalakrishnan2014, liu2003, noack2012, ramakrishnan2011, suri2011a, suri2012a, tete2010}.  Potentiation and depression rules are specific forms of spike-timing dependent plasticity (STDP) \cite{dan2004spike} rules.  STDP rules and their associated circuitry are extremely common in neuromorphic implementations for synapses \cite{afshar2015, ambrogio2013, ambrogio2016, ambrogio2016b, azghadi2011, azghadi2013, azghadi2013b, azghadi2014, bamford2012, bartolozzi2009, boegerhausen2003, cassidy2011, choi2011, dytckov2014, fusi2000, gopalakrishnan2014, gopalakrishnan2015, gopalakrishnan2015a, hayashi2007, hindo2014, indiveri2002, irizarry2014, kornijcuk2014, liu2008, mayr2010, mayr2013, meador1991, mitra2010, narasimman2016, noack2010, rachmuth2011, ramachandran2014, ramakrishnan2011, saeki2009, shahim2015, smith2014, srinivasan2016b, sumislawska2016, suri2012, wu2012e}.  More information on STDP as a learning algorithm and its implementations in neuromorphic systems is provided in Section \ref{sec:algorithms}.  Synaptic responses can also be relatively complex in neuromorphic systems.  Some neuromorphic synapse implementations focus on synaptic dynamics, such as the shape of the outgoing spike from the synapse or the post-synaptic potential \cite{chen2006, chen2006a, chen2008, ghani2011, ghani2012}.  Synapses in spiking neuromorphic systems have also been used as homeostasis mechanisms to stabilize the network's activity, which can be an issue in spiking neural network systems \cite{bartolozzi2006, liu2002}. 


\begin{figure*}[!t]
\centering
\includegraphics[width=0.8\textwidth]{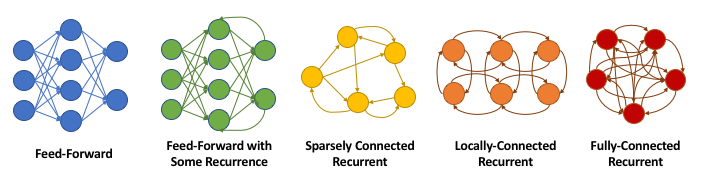}
\caption{Different network topologies that might be desired for neuromorphic systems. Determining the level of connectivity that is required for a neuromorphic implementation and then finding the appropriate hardware that can accommodate that level of connectivity is often a non-trivial exercise.}
\label{fig:network_topologies}
\end{figure*}


A variety of neuromorphic synaptic implementations for non-spiking neural networks have also been implemented.  These networks include feed-forward multi-layer networks \cite{cilingiroglu1990, chible2000, diorio1996, fuller2016, hasler1995, kim1992a, mcginnity1998}, winner-take-all \cite{yu2014a}, and convolutional neural networks \cite{vianello2017}.  A focus on different learning rules for synapses in artificial neural network-based neuromorphic systems is also common, as it is for STDP and potentiation and depression rules in spike-based neuromorphic systems.  Common learning rules in artificial neural network-based systems include Hebbian learning \cite{card1989, card1991, diorio1996, hasler1995} and least mean-square \cite{chiang1991, srinivasan2005}. Gaussian synapses have also been implemented in order to help with back-propagation learning rules \cite{choi1994, lau1997, lee1995a}. 

\subsection{Network Models}

\label{sec:network_models}

Network models describe how different neurons and synapses are connected and how they interact.  As may be intuited from the previous sections, there are a wide variety of neural network models that have been developed for neuromorphic systems.  Once again, they range from trying to replicate biological behavior closely to much more computationally-driven, non-spiking neural networks.  There are a variety of factors to consider when selecting a network model.  One of the factors is clearly biological inspiration and complexity of neuron and synapse models, as discussed in previous sections.  Another factor to consider is the topology of the network.  Figure \ref{fig:network_topologies} shows some examples of network topologies that might be used in various network models, including biologically-inspired networks and spiking networks.  Depending on the hardware chosen, the connectivity might be relatively restricted, which would restrict the topologies that can be realistically implemented. A third factor is the feasibility and applicability of existing training or learning algorithms for the chosen network model, which will be discussed in more detail in Section \ref{sec:algorithms}.  Finally, general applicability of that network model to a set of applications may also play a role in choosing the appropriate network model. 

There are a large variety of general spiking neural network implementations in hardware  \cite{annunziato1998, bartolozzi2004, bartolozzi2008, binas2016, carota2012, chicca2001, chicca2003a, chien2015, fieres2008, fusi1999, ghaderi2015, giulioni2011, glover1998, glover1999, glover2002, goldberg2001, goldberg2001a, grassia2011a, hafliger1997, hajtavs2000, huo2005, indiveri2000, indiveri2000a, indiveri2001b, indiveri2002a, indiveri2007, joubert2011, joubert2012, kanazawa2004, koickal2007, koickal2009, lazzaro1992, mahowald1997, mill2011, mitra2006, nease2013, neftci2013, oster2005, perez2014, pfeil2013a, pfeil2016, querlioz2013a, renaud2007, riis2004, rodrigues2014, saighi2010, schemmel2004, schemmel2007, schemmel2008, schemmel2012, schreiter2002, smith1998, tanaka2009, tovar2006, tovar2008, utagawa2007, vanschaik2001, vogelstein2002, wang2006, wang2009, wang2010, wijekoon2012a, yang2004, zhang2010, zhao2016b, ahmed2016a, ambroise2013, banerjee2015, dean2016, georgiou2006, hu2014, imam2013, kim2015c, kulkarni2015, nere2012, nere2013, roclin2013, schoenauer2000, schoenauer2002, seo2015, shen2016, zhang2015g, abinaya2015, afifi2011, arthur2006, arthur2007, asai2004a, azghadi2013a, azghadi2015, boi2016, bruderle2010, bruderle2011, chicca2004, chicca2006, chicca2007a, choudhary2012, corneil2012, corneil2012a,  corradi2014, corradi2015, cruz2012, folowosele2008, friedmann2013, gao2007, gao2008, hajtas2004, hammerstrom2009, han2006, han2006a, huo2012, hussain2012, indiveri2001a, indiveri2004, indiveri2006, indiveri2011b, kim2013a, linares2003, liu2004, mayr2014a, mayr2016a, mcdaid2008, meier2015, minkovich2012, mitra2007, mitra2007a, mitra2009, mitra2009a, moradi2011, moradi2014, mostafa2014, neftci2007, neftci2011, neftci2012a, noack2015, palma2013, petrovici2014, pfeil2013, qiao2015, renaud2010, schemmel2006, schemmel2010, schmuker2014, shi2015a, sinha2010, sinha2011, tapson2012, vogelstein2004, vogelstein2004a, vogelstein2007, wang2006b, wang2007, wang2014b, wang2014e, wang2014f, wang2014i, wang2015, wang2016g, xu2015, yu2012b, dahasert2012, farquhar2006, marr2015, mcginley2008, rocke2005, rocke2007, rocke2008, zhao2007, agis2007, caron2013, neil2014, ahn2013, ahn2015, ahn2015a, bako2009, bako2013, belhadj2009, bellis2004, bhuiyan2010a, blair2013, caron2011, cassidy2007, cassidy2008, cassidy2011b, chappet2014, cheung2012, cheung2015, cong2013, daffron2016, de2016a, dean2014, dean2016a, diaz2016, farsa2015, fox2012, garg2011, ghani2006, glackin2005a, glackin2009, glackin2009a, gomar2014, gomar2016, grassia2014, hafliger2001, harkin2008, harkin2009, hashimoto2010, hellmich2004, hellmich2004a, iakymchuk2012, iakymchuk2014, johnston2005, just2010, koziol2014, li2012, li2013f, makhlooghpour2016, maya2000, molin2015, moore2012, niu2012, nuno2011, nuno2012, pearson2005, pearson2005a, pearson2007, rice2009, rios2016, roggen2003, ros2003, rosado2011, rostro2011, rostro2012, sanchez2010, schrauwen2006, shayani2008a, shayani2008b, shayani2009, sheik2016, sofatzis2014, soleimani2012, upegui2005, wang2013, wang2014c, wang2014p, wang2014u, wang2015n, wu2015e, xicotencatl2003, yang2011, yang2011a, zuppicich2009}. These implementations utilize a variety of neuron models such as the various integrate-and-fire neurons listed above or the more biologically-plausible or biologically-inspired models. Spiking neural network implementations also typically include some form of STDP in the synapse implementation.  Spiking models have been popular in neuromorphic implementations in part because of their event-drive nature and improved energy efficiency relative to other systems.  As such, implementations of other neural network models have been created using spiking neuromorphic systems, including spiking feed-forward networks \cite{bofill2004, giulioni2016, bako2010, marti2016, nuno2009}, spiking recurrent networks \cite{badoni2006, camilleri2010, chicca2003, fusi2000a, giulioni2015, diehl2016}, spiking deep neural networks\cite{chevitarese2016, diehl2016a, esser2013, han2016b, indiveri2015c, cao2015, cerezuela2015, murphy2016, nurse2016, tsai2016, esser2016}, spiking deep belief networks \cite{stromatias2015a}, spiking Hebbian systems \cite{bofill2001, bofill2003, camilleri2007, giulioni2008, giulioni2008a, giulioni2009, gordon2002, hafliger1999a, sun2011}, spiking Hopfield networks or associative memories \cite{huayaney2011, shapero2013, ang2011}, spiking winner-take-all networks \cite{dungen2005, hafliger2007, liu2006, oster2004, oster2008, abrahamsen2004}, spiking probabilistic networks \cite{hsieh2013, hsieh2013a}, and spiking random neural networks \cite{abdelbaki2000}.   In these implementations a spiking neural network architecture in neuromorphic systems has been utilized for another neural network model type.  Typically, the training for these methods is done on the traditional neural network type (such as the feed-forward network), and then the resulting network solution has been adapted to fit the spiking neuromorphic implementation.  In these cases, the full properties of the spiking neural network may not be utilized.  

A popular biologically-inspired network model that is often implemented using spiking neural networks is central pattern generators (CPGs).  CPGs generate oscillatory motion, such as walking gaits or swimming motions in biological systems.  A common use of CPGs have been in robotic motion.  There are several neuromorphic systems that were built specifically to operate as CPGs \cite{donati2014, donati2016, ambroise2015, barron2013, joucla2016}, but CPGs are also often an application built on top of existing spiking neuromorphic systems, as is discussed in Section \ref{sec:applications}. 

The most popular implementation by far is feed-forward neural networks, including multi-layer perceptrons \cite{ abutalebi1998, akers1989, aksin2009, al2001, alhalabi1995, almeida1996, alspector1993, amaral2004, arima1992, bayraktarouglu1999, berg1996, bibyk1989, bibyk1990, bo1996, bo1997, bo1999, bollano1997, borgstrom1990, botha1992, bridges2005, cairns1994, calayir2015, carvajal2011, chang1996, chasta2012, cho1996, cho1998, cho1999, choi1991, choi1992, choi1993, choi1993a, choi1993b, choi1996, coggins1994, coggins1995, coggins1995a, coue1996, diotalevi2000, docheva2007, dolenko1993, dolenko1993a, dolenko1995, donald1993, duong1992, eberhardt1989, eberhardt1992, el1997, el2003, fakhraie1995, feltham1991, fisher1990, foruzandeh1999, foruzandeh1999a, furman1988, gatet2006, gatet2007, gatet2008, gatet2008a, gatet2008b, gatet2009, heruseto2009, hirotsu1993, hohmann2002, houselander1988, jabri1992, kakkar2009, lan1995, lansner1992, leong1992, lindblad1995, liu1999, liu1999a, lont1992, lont1993, lu2001, lu2001a, lu2001b, lu2002, lu2002a, lu2002b, lu2003, maeda1993, maeda1995, maliuk2010, maliuk2010a, maliuk2012, maliuk2014, maliuk2014a, masa1994, masmoudi1999, mestari2004, michel2004, milev2003, modi2006, montalvo1994, montalvo1997, montalvo1997a, morns1999, mundie1994, murray1992a, nosratinia1992a, oh1993, oh1994, pan2012, pinjare2009, richter2015, salam1990, satyanarayana1989, shimabukuro1989, soelberg1994, song1993, song1994, sun2002, tam1990, tam1992, tawel1993, thakur2015a, thakur2016, tombs1993, tsividis1987, valle1994, valle1996, van1992, walker1989, wang1993, wawryn2001, wawryn2001a, weller1990, withagen1994, wolpert1992, yildirim1996, yildiz2007, lee2014a, christiani2016, distante1990, distante1990a, fornaciari1994, hammerstrom1990, hammerstrom1991, islam2006, kim1992, kim1993, kumar1996, kung1989, larsson1996, myers1989, pakzad1993, pechanek1994, plaskonos1993, popescu2000, szabo2000, tang1997, tomlinson1990, tomlinson1990a, torbey1992, zhang1991, aihara1996, avellana1996, ayala2002, bermak1999, bermak2002, chang1992, distante1990b, distante1991, duranton1989, faiedh2004, joseph2010, kim2015g, orrey1991, tuazon1993, vincent1991, walker1988, alippi1991a, chung1992, cloutier1994, duranton1986, eguchi1991, kondo1994, madokoro2013, myers1993, oteki1993, saito1998, sato1993, tang1993, theeten1990, wang2006a, wawrzynek1992, yun2002, inigo1990, hikawa2001, hikawa2002, kim1995, kim1995a, al2000, barkan1990, binfet2001, bor1996, boser1991, camboni2001, cardarilli1994, cardarilli1995, corso1992, current1990, del2008, djahanshahi1996, djahanshahi1996a, djahanshahi1996b, djahanshahi1997, djahanshahi1997a, erkmen2013, fakhraie2004, fieres2004, franca1993, graf1990, jackel1987, jackel1990, jackel1990a, johnson1995, koosh2001, koosh2002, lee2006a, lehmann1996, likharev2011, liu2002a, liu2010a, maeda1999, masa1994a, maundy1991b, mirhassani2003, mirhassani2004, mirhassani2005, mirhassani2007, morie2000, nosratinia1992, pan2003, passos1993, sackinger1991, sackinger1992, sanchez2013, schemmel2004a, schmid1999a, tarassenko1991, van1990, watanabe1997, yang1999, yazdi1993, zatorre2006, burger2014, dong2014, emelyanov2015, emelyanov2016, manem2015, merkel2014c, such2015, dong2006, liu2009b, al2008, banuelos2003, baptista2015, bastos2006, blake1997, blake1998, bohrn2013, bonnici2006, botros1993, braga2012, brunelli2005, chujo2000, deng2014, deotale2014, dinu2007, dinu2010, dondon2014, dong2011, economou1994, ehkan2014, erdogan1992, ferreira2004, ferreira2007, granado2006, guccione1994, hariprasath2012, hasanien2011, hikawa1995, himavathi2007, hoelzle2009, hoffman2006, izeboudjen1999, jeyanthi2014, joost2012, jung2007, kim2004, krcma2015, krips2002, kung2002, laudani2014, lotrivc2012, lozito2014, makwana2013, mand2012, mohamad2012, mohammed2013, muthuramalingam2008, noory2003, oniga2004, oniga2007, oniga2008, oniga2009, orlowska2011, ortega2014, perez1996, qi2014, raeisi2006, restrepo2000, sahin2006, salapura1994, salem2005, savran2003, shaari2008, shah2012, shreejith2016, soares2006, sonowal2012, wang2015b, wang2015g, wang2015m, won2007, youssef2012, zhang2005, zhang2006, ahn2013a, aliaga2009, alizadeh2008, biradar2015, cavuslu2011, domingos2005, eldredge1994, eldredge1994a, gadea2000, girau1996, girau2001, girones2005, hikawa1999, izeboudjen2007, moreno2009, moussa2006, nichols2002, ortega2015, pandya2005, pinjare2012, ruan2005, sangeetha2013, savich2007, shoushan2010, sun2012, acosta2001, ago2013, aliaga2008, ann2016, bahoura2011, bahoura2011a, benrekia2009, beuchat1998, beuchat1998a, beuchat1999, botros1994, braga2010, canas2006, carvalho2005, chalhoub2006, da2009, denby2003, ferreira2010, ferrer2004, gatet2009a, gomperts2010, gomperts2011, gorgon2006, horita2015, jin2011, johnston2005, khan2006, khodja2010, kim1998, kyoung2006, latino2009, lee2006, leon1999, lin2010, lotrivc2011, nedjah2008, nedjah2009, nedjah2012, nedjah2014, ortigosa2003, ortigosa2003a, ortigosa2006, ortigosa2006a, ozdemir2011, panicker2012, pasero2004, patra2006, perez2014a, qinruo2003, rani2007, rezvani2012, skodzik2013, syiam2003, taright1998, tatikonda2008, vitabile2005, vskoda2011, wolf2001, wolf2001a, yu2006, zhu1999}.  Extreme learning machines are a special case of feed-forward neural networks, where a number of the weights in the network are randomly set and never updated based on a learning or training algorithm; there have been several neuromorphic implementations of extreme learning machines \cite{basu2013, merkel2014, merkel2014b, suri2015b, yao2013, decherchi2012}.  Another special case of feed-forward neural networks are multi-layer perceptrons with delay, and those have also been implemented in neuromorphic systems \cite{gatt2000, van1994, bahoura2012, bahoura2014, ntoune2012}.  Probabilistic neural networks, yet another special-case of feed-forward neural networks that have a particular type of functionality that is related to Bayesian calculations, have several neuromorphic implementations \cite{woodburn2000, serb2016, aibe2002, aibe2004, bu2004, figueiredo1998, minchin1999, zhou2010, zhu2010}.  Single-layer feed-forward networks that utilize radial basis functions as the activation function of the neurons have also been used in neuromorphic implementations \cite{dogaru1996, sitte2007, verleysen1994, yildirim1996, maffezzoni1994, suri2015, erkmen2013, fakhraie2004, zhuang2007, halgamuge1994, johnston2005, kim2008, kim2015f, porrmann2002}. In recent years, with the rise of deep learning, convolutional neural networks have also seen several implementations in neuromorphic systems \cite{chakradhar2010, chen2015c, chen2016c, conti2015, kim2016b, nomura2004, nomura2005, kang2015, korekado2003, garbin2014, garbin2015, garbin2015a, chung2016, farabet2009, farabet2011, li2016b, motamedi2016, qiao2016, qiu2016a, shin2016, suda2016, zhang2015d}.

Recurrent neural networks are those that allow for cycles in the network, and they can have differing levels of connectivity, including all-to-all connections. Non-spiking recurrent neural networks have also been implemented in neuromorphic systems \cite{boddhu2006, brownlow1991, cauwenberghs1994, cauwenberghs1996, cauwenberghs1996a, fisher1991, gallagher2000, gallagher2001, gallagher2005, gallagher2008, hasler1991, kothapalli2005, lansner1993, morie1994, salam1989, salam1989a, salam1991, schemmel2001, thakoor1991, boser1991, graf1990, ota1996, van1990, girau2000, gupta2009, li2015h, lin2009, maeda2005}.  Reservoir computing models, including liquid state machines, have become popular in neuromorphic systems \cite{polepalli2016, polepalli2016a, petre2016, kudithipudi2015, schrauwen2008, wang2016l, yi2016}.  In reservoir computing models, recurrent neural networks are utilized as a ``reservoir", and outputs of the reservoir are fed into simple feed-forward networks.  Both spiking and non-spiking implementations exist.  Winner-take-all networks, which utilize recurrent inhibitory connections to force a single output, have also been implemented in neuromorphic systems \cite{hylander1993, neftci2010, neftci2010a}.  Hopfield networks were especially common in earlier neuromorphic implementations, as is consistent with neural network research at that time \cite{kamio1997, khachab1989, lee1990a, linares1992, linares1992a, paulos1988, verleysen1989, yang1994, alla1991, asari1994, blayo1989, blayo1989a, fornaciari1994a, johannet1992, kumar1996, lehmann1993, masaki1990, pechanek1994, van1989, weinfeld1989, wike1990, yasunaga1993, tomberg1989, tomberg1990, aibara1991, boser1991, chen1996, hansen1989, hollis1990, lee1990b, lee1991, maundy1990, moopenn1990, murray1988, murray1991, tank1986, abramson1998, likharev2003a, atencia2007, boumeridja2005, de2002a, gschwind1996, saif2006, stepanova2007, varma2002, wakamura2003}, but there are also more recent implementations \cite{likharev2011,  duan2015, guo2015b, hu2015, liu2012, liu2013b, liu2015d, clemente2016, harmanani2010, mansour2011, sousa2014, srinivasulu2012}. Similarly, associative memory based implementations were also significantly more popular in earlier neuromorphic implementations \cite{andreou1989, andreou1990, boahen1989, boahen1989a, howard1987, kaulmann2005, linares1993, maundy1991a, mccarley1995, saeki2011, hasan1995, graf1987a, graf1988, heittmann2002, ruckert1988, ruckert1991, wu1996, pershin2010, wang2015j, hammerstrom2003, leiner2008, li2013e, porrmann2002, reay1994}. 

Stochastic neural networks, which introduce a notion of randomness into the processing of a network, have been utilized in neuromorphic systems as well \cite{neftci2016, neftci2016a, torralba1995, fang1990, bade1994, li2004, nedjah2003, nedjah2003a, nedjah2007, van1993}.  A special case of stochastic neural networks, Boltzmann machines, have also been popular in neuromorphic systems.  The general Boltzmann machine was utilized in neuromorphic systems primarily in the early 1990's \cite{ jayakumar1992, madani1991, pujol1994, schneider1991b, schneider1993, arima1991, arima1991a, sato1992}, but it has seen occasional implementations in more recent publications \cite{bojnordi2016, chen2003, lu2007, lu2009}.  A more common use of the Boltzmann model is the restricted Boltzmann machine, because the training time is significantly reduced when compared with a general Boltzmann machine.  As such, there are several implementations of the restricted Boltzmann machine in neuromorphic implementations \cite{knag2016, pedroni2016, chen2003, lu2007, lu2009, rafique2016, sheri2015, suri2015a, kim2009, kim2014b, le2010}.  Restricted Boltzmann machines are an integral component to deep belief networks, which have become more common with increased interest in deep learning and have been utilized in neuromorphic implementations \cite{das2015, ahn2014, sanni2015}.

Neural network models that focus on unsupervised learning rules have also been popular in neuromorphic implementations, beyond the STDP rules implemented in most spiking neural network systems.  Hebbian learning mechanisms, of which STDP is one type in spiking neural networks, are common in non-spiking implementations of neuromorphic networks \cite{card1994, cauwenberghs1991, schneider1991, schneider1991a, schneider1991c, shima1992, del1998, likharev2003, zaman1994, aggarwal2012, ascoli2014, cantley2011, cantley2012, kubendran2012, wang2014q, wen2013b, ziegler2013}.  Self-organizing maps are another form of artificial neural network that utilize unsupervised learning rules, and they have been utilized in neuromorphic implementations \cite{andreou1989, dlugosz2010, mann1989, maundy1991, sheu1992, cambio2003, dlugosz2011a, kumar1996, lehmann1991, marchese2015, mauduit1992, rajah2004, rodriguez2015, fang1992, mann1988, ben2004, brassai2014, porrmann2002, porrmann2002a, tamukoh2010}.  More discussion on unsupervised learning methods such as Hebbian learning or STDP is provided in Section \ref{sec:algorithms}. 

The visual system has been a common inspiration for artificial neural network types, including convolutional neural networks.  Two other visual system-inspired models, cellular neural networks \cite{chua1988cellular} and pulse-coupled neural networks \cite{wang2010pulse}, have been utilized in neuromorphic systems.  In particular, cellular neural networks were common in early neuromorphic implementations \cite{cruz1991, cruz1998, dalla1993, halonen1991, harrer1992, kinget1994, kinget1995, rodriguez1993, yang1990, salerno1999} and have recently seen a resurgence \cite{arena2006, ascoli2016, corinto2014, duan2014a, guo2014, qin2015, martinez2003, rak2009}, whereas pulse-coupled networks were popular in the early 2000's \cite{chen2007, chen2010a, ota1999, xiong2010, matolin2004, matolin2004a, torikai2005, ehrlich2007, vega2006}.  

\begin{figure}[!t]
\centering
\includegraphics[width=0.5\textwidth]{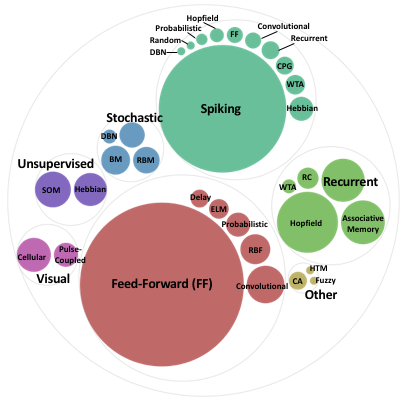}
\caption{A breakdown of network models in neuromorphic implementations, grouped by overall type and sized to reflect the number of associates papers.}
\label{fig:network_model_pie_chart}
\end{figure}

\begin{figure*}[!t]
\centering
\includegraphics[width=0.7\textwidth]{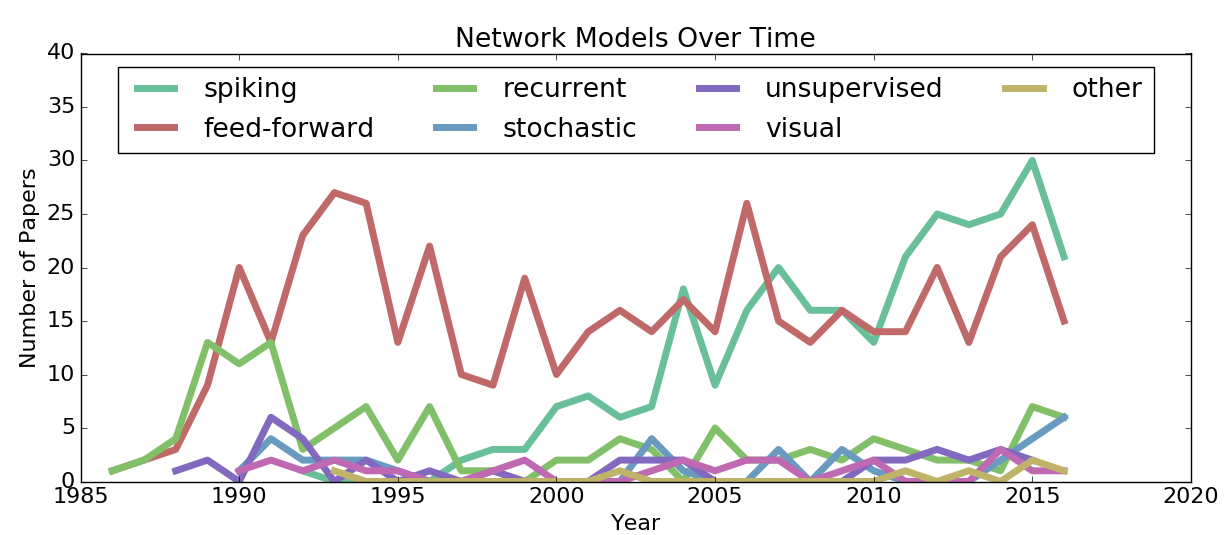}
\caption{An overview of how models for neuromorphic implementations have changed over time, in terms of the number of papers published per year.}
\label{fig:network_models_over_time}
\end{figure*}

Other, less common neural network and neural network-adjacent models implemented in neuromorphic systems include cellular automata \cite{secco2015, de2002, isobe2015, matsubara2011, matsubara2013}, fuzzy neural networks \cite{kuo1993}, which combine fuzzy logic and artificial neural networks, and hierarchical temporal memory \cite{ibrayev2016}, a network model introduced by Hawkins in \cite{hawkins2007intelligence}. 

%
%
%

Figure \ref{fig:network_model_pie_chart} gives an overview of the network models implemented in neuromorphic systems.  Figure \ref{fig:network_models_over_time} shows how some of the most frequently used models in neuromorphic implementations have evolved over time.  As can be seen in the figures, spiking and feed-forward implementations are by far the most common, with spiking implementations seeing a rise in the last decade.  General feed-forward networks had begun to taper off, but the popularity and success of convolutional neural networks in deep learning has increased in activity in the last five years.  

\subsection{Summary and Discussion}

In terms of model selection for neuromorphic implementations, it is clear that there are a wide variety of options, and much of the ground of potential biological and artificial neural network models has been tread at least once by previous work.  The choice of model will be heavily dependent on the intent of the neuromorphic system.  With projects whose goal it is to produce useful results for neuroscience, the models usually err on the side of biologically-plausible or at least biologically-inspired.  For systems that have been moved to hardware for a particular application, such as image processing on a remote sensor or autonomous robots, more artificial neural network-like systems that have proven capabilities in those arenas may be most applicable.  It is also the case that the model is chosen or adapted to fit within some particular hardware characteristics (e.g., selecting models that utilize STDP for memristors), or that the model is chosen for efficiency's sake, as is often the case for event-driven spiking neural network systems.  On the whole, it is clear that most neural network models have, at some point in their history, been implemented in hardware.  

\section{Algorithms and Learning}
\label{sec:algorithms}

Some of the major open questions for neuromorphic systems revolve around algorithms.  The chosen neuron, synapse, and network models have an impact on the algorithm chosen, as certain algorithms are specific to certain network topologies, neuron models, or other network model characteristics.  Beyond that, a second issue is whether training or learning for a system should be implemented on chip or if networks should be trained off-chip and then transferred to the neuromorphic implementation.  A third issue is whether the algorithms should be on-line and unsupervised (in which case they would necessarily need to be on-chip), whether off-line, supervised methods are sufficient, or whether a combination of the two should be utilized.  One of the key reasons neuromorphic systems are seen as a popular post-Moore's law era complementary architecture is their potential for on-line learning; however, even the most well-funded neuromorphic systems struggle to develop algorithms for programming their hardware, either in an off-line or on-line way.  In this section, we focus primarily on on-chip algorithms, chip-in-the-loop algorithms, and algorithms that are tailored directly for the hardware implementation. 

\subsection{Supervised Learning}


The most commonly utilized algorithm for programming neuromorphic systems is back-propagation.  Back-propagation is a supervised learning method, and is not typically thought of as an on-line method.  Back-propagation and its many variations can be used to program feed-forward neural networks, recurrent neural networks (usually back-propagation through time), spiking neural networks (where often feed-forward neural networks are adapted to spiking systems), and convolutional neural networks.  The simplest possible approach is to utilize back-propagation off-line on a traditional host machine, as there are many available software implementations that have been highly optimized.  We omit citing these approaches, as they typically utilize basic back-propagation, and that topic has been covered extensively in the neural network literature \cite{haykin2004neural}.  However, there are also a large variety of implementations for on-chip back-propagation in neuromorphic systems \cite{ahn2013a, alhalabi1995, aliaga2008, aliaga2009, alippi1991a, alizadeh2008, ayala2002, bahoura2011, bahoura2011a, bahoura2012, bastos2006, bayraktarouglu1999, berg1996, beuchat1998, beuchat1998a, beuchat1999, bibyk1989, biradar2015, bo1999, botros1993, cavuslu2011, chang1992, chasta2012, cho1996, cho1998, cho1999, choi1992, choi1993, choi1996, chung1992, cloutier1994, damak2006, distante1991, dolenko1993, dolenko1993a, dolenko1995, domingos2005, duong1992, duranton1989, eberhardt1989, eldredge1994, eldredge1994a, furman1988, gadea2000, girau1996, girau2001, girones2005, gomperts2010, gomperts2011, hasan2014, hikawa1995, hikawa1999, hikawa2003, hikawa2003a, hollis1994, ishii1992, izeboudjen2007, jung2007, kim2015f, kondo1994, kondo1995, krid2006, kumar1996, liu2010a, lu2001, lu2001a, lu2001b, lu2002, lu2002a, lu2002b, lu2003, maundy1991b, moreno2009, morie1994, moussa2006, myers1993, nichols2002, ntoune2012, nuno2009, orrey1991, ortega2015, oteki1993, pandya2005, pechanek1994, perez2014a, pinjare2009, pinjare2012, prezioso2015a, ruan2005, sangeetha2013, sato1993, savich2007, shoushan2010, soelberg1994, song1993, song1994, soudry2015, sun2012, tam1990, tang1993, theeten1990, valle1996, van1990, vincent1991, wang1993, wang2006a, wawrzynek1992, withagen1994, wolpert1992, yasunaga1993, yu2006, yun2002, zamanidoost2015 }.  There have been several works that adapt or tailor the back-propagation method to their particular hardware implementation, such as coping with memristive characteristics of synapses \cite{bollano1997, lont1992, negrov2016, neil2016, ueda2014}.  Other gradient descent-based optimization methods have also been implemented on neuromorphic systems for training, and they tend to be variations of back-propagation that have been adapted or simplified in some way \cite{bor1996, carvajal2011, choi1991, eguchi1991, gatt2000, hussain2013, johannet1992, li2014d, milev2003, mirhassani2003, murray1992a, nair2015, oh1993, oh1994, rosenthal2016, salam1990}.  Back-propagation methods have also been developed in chip-in-the-loop training methods \cite{cardarilli1994, lindblad1995, maliuk2012, tawel1993, yang1999}; in this case, most of the learning takes place on a host machine or off-chip, but the evaluation of the solution network is done on the chip.  These methods can help to take into account some of the device's characteristics, such as component variation.  


There are a variety of issues associated with back-propagation, including that it is relatively restrictive on the type of neuron models, networks models, and network topologies that can be utilized in an efficient way.  It can also be difficult or costly to implement in hardware.   Other approaches for on-chip supervised weight training have been utilized.  These approaches include the least-mean-squares algorithm \cite{christiani2016, merkel2014, merkel2014b, walker1988}, weight perturbation \cite{alspector1993, cairns1994, cauwenberghs1994, cauwenberghs1996, cauwenberghs1996a, diotalevi2000, foruzandeh1999, jabri1992, koosh2001, koosh2002, lin2009, maeda1993, maeda1995, maeda1999, maeda2003, maeda2005, michel2004, mirhassani2004, mirhassani2005, mirhassani2007, modi2006, montalvo1997, montalvo1997a, mundie1994, schmid1999a, tsividis1987, wakamura2003}, training specifically for convolutional neural networks \cite{fieres2006, fieres2006a} and others \cite{bennett2016, chabi2011, chabi2014a, chabi2015, chabi2015a, decherchi2012, emelyanov2015, emelyanov2016, hashimoto2010, hu2014a, inigo1990, lewis2000, liu2013a, merkel2013, morns1999, petridis1995, qiu2014, qiu2015, schmid1999, sinha2010, taha2014, woodburn1994, zhuang2007}.   Other on-chip supervised learning mechanisms are built for particular model types, such as Boltzmann machines, restricted Boltzmann machines, or deep belief networks \cite{arima1992, jayakumar1992, lee1991, madani1991, schneider1993, chen2003, lu2007, sheri2015} and hierarchical temporal memory \cite{ibrayev2016}. 

A set of nature-based or evolution-inspired algorithms have also been also been implemented for hardware.  These implementations are popular because they do not rely on particular characteristics of a model to be utilized, and off-chip methods can easily utilize the hardware implementations in the loop.  They can also be used to optimize within the characteristics and peculiarities of a particular hardware implementation (or even the characteristics and peculiarities of a particular hardware device instance).  Off-chip nature-based implementations include differential evolution \cite{buhry2009, buhry2009a, buhry2012, grassia2011}, evolutionary or genetic algorithms \cite{boddhu2006, buhry2008, carlson2013, carlson2014, carlson2015, dean2016a, gallagher2000, gallagher2001, gallagher2005, gallagher2008, hohmann2002, howard2011, howard2012, howard2014, howard2014a, howard2015, maher2006, maliuk2010, mcginley2008, orchard2008, rocke2005, rocke2007, rocke2008, schemmel2001, schuman2016, schuman2016a, zuppicich2009}, and particle swarm optimization \cite{braendler2002a}.  We explicitly specify these off-chip methods because all of the nature-based implementations rely on evaluations of a current network solution and can utilize the chip during the training process (as a chip-in-the-loop method).  There have also been a variety of implementations that include the training mechanisms on the hardware itself or in a companion hardware implementation, including both evolutionary/genetic algorithms \cite{al2001, amaral2004, cawley2010, cawley2011, fe2013, harkin2008, low2006, merchant2006, merchant2006a, merchant2008, merchant2010, morgan2009, nambiar2014, nirmaladevi2015, shayani2008a, shayani2008b, shayani2009, upegui2005, upegui2006} and particle swarm optimization \cite{bezborah2012, cavuslu2012, farmahini2008, lin2008}.

 \begin{figure}[!t]
\centering
\includegraphics[width=0.45\textwidth]{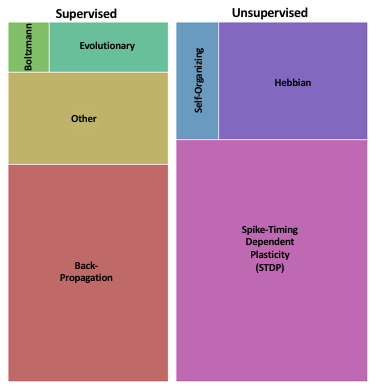}
\caption{An overview of on-chip training/learning algorithms.  The size of the box corresponds to the number of papers in that category.}
\label{fig:on_chip_algorithms}
\end{figure}

\begin{table*}[]
\centering
\caption{Algorithms Pros and Cons}
\label{tab:algorithms_pros_and_cons}
\begin{tabular}{|l|c|c|c|c|c|c|c|}
\hline
\textbf{Algorithm Class} & \textbf{\begin{tabular}[c]{@{}c@{}}Any \\ Model\end{tabular}} & \textbf{\begin{tabular}[c]{@{}c@{}}Device \\ Quirks\end{tabular}} & \textbf{\begin{tabular}[c]{@{}c@{}}Complex to \\ Implement\end{tabular}} & \textbf{On-Line} & \textbf{\begin{tabular}[c]{@{}c@{}}Fast Time \\ to Solution\end{tabular}} & \textbf{\begin{tabular}[c]{@{}c@{}}Demonstrated  \\ Broad Applicability\end{tabular}} & \textbf{\begin{tabular}[c]{@{}c@{}}Biologically-Inspired \\ or Plausible\end{tabular}} \\ \hline
Back-Propagation         & No                                                            & No                                                                & Yes                                                                      & No               & Yes                                                                       & Yes                                                                                   & No                                                                                     \\ \hline
Evolutionary             & Yes                                                           & Yes                                                               & No                                                                       & No               & No                                                                        & Yes                                                                                   & Maybe                                                                                   \\ \hline
Hebbian                  & No                                                            & Yes                                                               & No                                                                       & Yes              & Maybe                                                                     & No                                                                                    & Yes                                                                                    \\ \hline
STDP                     & No                                                            & Yes                                                               & Maybe                                                                    & Yes              & Maybe                                                                     & No                                                                                    & Yes                                                                                    \\ \hline
\end{tabular}
\end{table*}

\subsection{Unsupervised Learning}
There have been several implementations of on-chip, on-line, unsupervised training mechanisms in neuromorphic systems.  These self-learning training algorithms will almost certainly be necessary to realize the full potential of neuromorphic implementations.  Some early neuromorphic implementations of unsupervised learning were based on self-organizing maps or self-organizing learning rules \cite{arima1991, arima1991a, ben2004, kumar1996, mann1988, mann1989, matolin2004, matolin2004a, maundy1991, ota1999, porrmann2002, porrmann2002a, rajah2004, sheu1992, yu2006}, though there have been a few implementations in more recent years \cite{dlugosz2011a, rodriguez2015, tamukoh2010, xiong2010}.  Hebbian-type learning rules, which encompass a broad variety of rules, have been very popular as on-line mechanisms for neuromorphic systems, and there are variations that encompass both supervised and unsupervised learning \cite{annunziato1998, ascoli2014, badoni2006, bako2010, bako2013, bofill2001, bofill2003, camilleri2007, cantley2011, cantley2012, card1994, cho1996, de2002a, del1998, donald1993, duan2015, gao2007, gao2008, giulioni2008a, giulioni2009, giulioni2011, giulioni2015, gordon2002, gschwind1996, hafliger1997, hafliger1999a, hafliger2007, hajtavs2000, hasler1992, hsieh2013, hu2012, hu2015, jackson2016, kubendran2012, li2013e, likharev2003, linares1992a, linares1993, liu2012, liu2013b, liu2015d, mahowald1997, mansour2011, merrikh2014, neftci2016a, oniga2008, pfeil2013a, porrmann2002, rachmuth2003, riis2004, rossmann1997, schafer1999, schneider1991, schneider1991a, schneider1991b, schneider1991c, soltiz2012, stepanova2007, sun2011, theeten1990, tomberg1991, wang2014q, weinfeld1989, wen2013b, yang2004, zaman1994, ziegler2013 }.    Finally, perhaps the most popular on-line, unsupervised learning mechanism in neuromorphic systems is spike-timing dependent plasticity \cite{acciarito2016, afifi2009, afifi2009a, afifi2011, ahmed2016a, arthur2004, arthur2006, azghadi2013a, azghadi2015, azghadi2016, babacan2017, belhadj2009, bennett2016, berdan2016, bill2014, bofill2004, boi2016, brink2013, bruderle2011, cai2012, cai2015, cantley2011a, cassidy2007, chan2012, chen2014a, chen2015b, chen2016b, cheung2015, chicca2001, chicca2003, chicca2003a, chu2015, covi2015, covi2016a, cruz2012, cruz2013, dai2014, desalvo2015, du2015a, ebong2010, ebong2011, ebong2012, friedmann2013, fusi2000a, gale2012, gamrat2015, garbin2015, garg2011, giulioni2008, glackin2005a, glackin2009, grassia2011a, he2014, howard2012, howard2014, howard2014a, howard2015, hu2013c, hu2014, hu2016b, huayaney2011, huayaney2016, huayaney2016a, huo2005, huo2012, indiveri2002a, indiveri2004, indiveri2006, indiveri2007, indiveri2007a, indiveri2011b, indiveri2015c, jo2010, kaneko2013, kaneko2014, kavehei2011, kim2012a, kim2015a, koickal2007, koickal2009, kulkarni2015, lecerf2013, li2015c, li2015f, li2016c, likharev2011, linares2009, mahalanabis2016, mandal2013, mayr2016a, mazumder2016, meier2015, meng2011, mitra2006, mitra2007, mitra2007a, mitra2009, mitra2009a, moon2014, mostafa2014, mostafa2015, mostafa2016, naous2016a, nease2013, nere2012, nere2013, noack2015, pantazi2016, park2012, park2013a, park2016, payvand2014, payvand2015, perez2010, petrovici2014, pfeil2013, prezioso2016, prezioso2016a, prezioso2016b, prezioso2016d, qiao2015, querlioz2012, ren2013, renaud2007, renaud2010, roclin2013, rosado2011, saeki2011, saighi2010, saighi2010a, sanchez2010, schemmel2004, schemmel2006, schemmel2007, schemmel2008, schemmel2010, schmuker2014, seo2011a, seo2015, serb2016, serrano2012, serrano2014, shahsavari2016, sheik2012a, sheri2014, singha2014, snider2008, sofatzis2014, subramaniam2013, suri2013, tanaka2009, tovar2006, tovar2008, vogelstein2002, wang2013, wang2014b, wang2014e, wang2014f, wang2014j, wang2014r, wang2015, wang2016g, werner2016, werner2016a, wijekoon2012, wijekoon2012a, wozniak2016, wu2015, wu2015d, wu2015e, wu2015g, yang2013, zhang2010, zhang2014, zheng2015, zou2006, zou2006a}, which is a form of Hebbian-like learning that has been observed in real biological systems \cite{sjostrom2010spike}.  The rule for STDP is generally that if a pre-synaptic neuron fires shortly before (after) the post-synaptic neuron, the synapse's weight will be increased (decreased) and the less time between the fires, the higher the magnitude of the change. There are also custom circuits for depression \cite{asai2004a, kanazawa2003, kanazawa2004, kang2015a, liu2004, mill2011} and potentiation \cite{schultz1995} in synapses in more biologically-inspired implementations.  It is worth noting that, especially for STDP, wide applicability to a set of applications has not been fully demonstrated.

\subsection{Summary and Discussion}

Spiking neural network-based neuromorphic systems have been popular for several reasons, including the power and/or energy efficiency of their event-driven computation and their closer biological inspiration relative to artificial neural networks in general.  Though there have been proposed methods for training spiking neural networks that usually utilize STDP learning rules for synaptic weight updates, we believe that the full capabilities of spiking neuromorphic systems have not yet been realized by training and learning mechanisms.  As noted in Section \ref{sec:network_models}, spiking neuromorphic systems have been frequently utilized for non-spiking network models. These models are attractive because we typically know how to train them and how to best utilize them, which gives a set of applications for spiking neuromorphic systems.  However, we cannot rely on these existing models to realize the full potential of neuromorphic systems.  As such, the neuromorphic computing community needs to develop algorithms for spiking neural network systems that can fully realize the characteristics and capabilities of those systems.  This will require a paradigm shift in the way we think about training and learning.  In particular, we need to understand how to best utilize the hardware itself in training and learning, as neuromorphic hardware systems will likely allow us to explore larger-scale spiking neural networks in a more computationally and resource efficient way than is possible on traditional von Neumann architectures.  

An overview of on-chip learning algorithms is given in Figure \ref{fig:on_chip_algorithms}.  When choosing the appropriate algorithm for a neuromorphic implementation, one must consider several factors: (1) the chosen model, (2) the chosen material or device type, (3) whether learning should be on-chip, (4) whether learning should be on-line, (5) how fast learning or training needs to take place, (6) how successful or broadly applicable the results will be, and (7) whether the learning should be biologically-inspired or biologically-plausible.  Some of these factors for various algorithms are considered in Table \ref{tab:algorithms_pros_and_cons}.  For example, back-propagation is a tried and true algorithm, has been applied to a wide variety of applications and can be relatively fast to converge to a solution.  However, if a device is particularly restrictive (e.g., in terms of connectivity or weight resolution) or has a variety of other quirks, then back-propagation requires significant adaptation to work correctly and may take significantly longer to converge.  Back-propagation is also very restrictive in terms of the types of models on which it can operate.  We contrast back-propagation with evolutionary-based methods, which can work with a variety of models, devices, and applications.  Evolutionary methods can also be relatively easier to implement than more analytic approaches for different neuromorphic systems.  However, they can be  slow to converge for complex models or applications.  Additionally, both back-propagation and evolutionary methods require feedback, i.e., they are supervised algorithms.  Both Hebbian learning and STDP methods can be either supervised or unsupervised; they are also biologically-inspired and biologically-plausible, making them attractive to developers who are building biologically-inspired devices.  The downside to choosing Hebbian learning or STDP is that they have not been demonstrated to be widely applicable.

There is still a significant amount of work to be done within the field of algorithms for neuromorphic systems.  As can be seen in Figure \ref{fig:network_models_over_time}, spiking network models are on the rise.  Currently, STDP is the most commonly used algorithm proposed for training spiking systems, and many spiking systems in the literature do not specify a learning or training rule at all.  It is worth noting that algorithms such as back-propagation and the associated network models were developed with the von Neumann architecture in mind.  Moving forward for neuromorphic systems, algorithm developers need to take into account the devices themselves and have an understanding of how these devices can be utilized most effectively for both learning and training.  Moreover, algorithm developers need to work with hardware developers to discover what can be done to integrate training and learning directly into future neuromorphic devices, and to work with neuroscientists in understanding how learning is accomplished in biological systems. 


\section{Hardware}
\label{sec:hardware}

Here we divide hardware implementations of neuromorphic implementations into three major categories:
digital, analog, and mixed analog/digital platforms.  These are examined at a high-level with some of the more exotic device-level components utilized in neuromorphic systems explored in greater depth.  For the purposes of this survey, we maintain a high-level view of the neuromorphic system hardware considered.

\subsection{High-Level}

\begin{figure}[!t]
\centering
\includegraphics[width=0.45\textwidth]{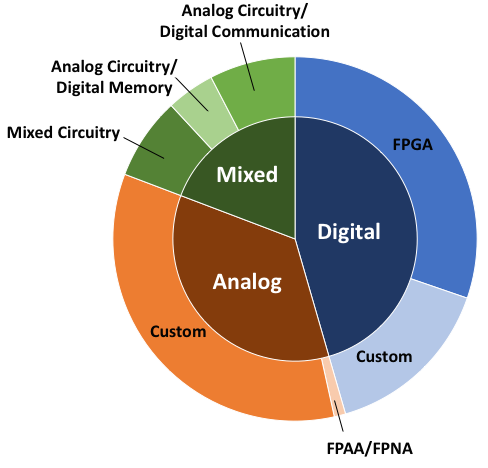}
\caption{An overview of hardware implementations in neuromorphic computing.  These implementations are relatively basic hardware implementations and do not contain the more unusual device components discussed in Section \ref{sec:devicelevel}.}
\label{fig:hardware_breakdown_piechart}
\end{figure}

There have been many proposed taxonomies for neuromorphic hardware systems \cite{izeboudjen2014}, but most of those taxonomies divide the hardware systems at a high-level into analog, digital or mixed analog/digital implementations.  Before diving into the neuromorphic systems themselves, it is worthwhile to note the major characteristics of analog and digital systems and how they relate to neuromorphic systems.  Analog systems utilize native physical characteristics of electronic devices as part of the computation of the system, while digital systems tend to rely on Boolean logic-based gates, such as AND, OR, and NOT, for building computation.  The biological brain is an analog system and relies on physical properties for computation and not on Boolean logic.  Many of the computations in neuromorphic hardware lend themselves to the sorts of operations that analog systems naturally perform.  Digital systems rely on discrete values while analog systems deal with continuous values.  Digital systems are usually (but not always) synchronous or clock-based, while analog systems are usually (but not always) asynchronous; in neuromorphic, however, this rule of thumb is often not true, as even the digital systems tend to be event-driven and analog systems sometimes employ clocks for synchronization.  Analog systems tend to be significantly more noisy than digital systems; however, there have been some arguments that because neural networks can be robust to noise and faults, they may be ideal candidates for analog implementation \cite{indiveri2002b}.  Figure \ref{fig:hardware_breakdown_piechart} gives an overall summary breakdown of high-level breakdown of different neuromorphic hardware implementations. 

\subsubsection{Digital}
\label{sec:digital}
Two broad categories of digital systems are addressed here.  The first is field programmable gate arrays or FPGAs.  FPGAs have been frequently  utilized  in neuromorphic systems \cite{hammerstrom2003, leiner2008, li2013e, porrmann2002, reay1994, jimenez2016, jones2000, thakur2015, trujillo2012, bailey2011, belhadj2008, berzish2016, beuler2012, bonabi2012, bonabi2014, cassidy2013, dangelo2015, glackin2005, graas2004, leung2008, machado2014, majani2012, mak2005, mak2006, moctezuma2013, moctezuma2015, mokhtar2008, nazari2014, nazari2015, nazari2015a, nazari2015b, nouri2015, pourhaj2010, pourhaj2010a, prieto2005, rossmann1996, shayani2008, shi2006, smaragdos2014, thomas2009, weinstein2005, weinstein2006, weinstein2007, yang2015b, zhang2009, ambroise2015, barron2013, joucla2016, de2002, isobe2015, matsubara2011, matsubara2013, martinez2003, rak2009, farabet2009, farabet2011, li2016b, motamedi2016, qiao2016, qiu2016a, shin2016, suda2016, zhang2015d, ahn2014, sanni2015, decherchi2012, van2010, vourkas2016, agis2007, caron2013, neil2014, andreou2016a, azhar2002, braendler2002, brauer1999, byungik2012, krcma2015a, li2016e, moreno1999, xiaobin2003, al2008, banuelos2003, baptista2015, bastos2006, blake1997, blake1998, bohrn2013, bonnici2006, botros1993, braga2012, brunelli2005, chujo2000, deng2014, deotale2014, dinu2007, dinu2010, dondon2014, dong2011, economou1994, ehkan2014, erdogan1992, ferreira2004, ferreira2007, granado2006, guccione1994, hariprasath2012, hasanien2011, hikawa1995, himavathi2007, hoelzle2009, hoffman2006, izeboudjen1999, jeyanthi2014, joost2012, jung2007, kim2004, krcma2015, krips2002, kung2002, laudani2014, lotrivc2012, lozito2014, makwana2013, mand2012, mohamad2012, mohammed2013, muthuramalingam2008, noory2003, oniga2004, oniga2007, oniga2008, oniga2009, orlowska2011, ortega2014, perez1996, qi2014, raeisi2006, restrepo2000, sahin2006, salapura1994, salem2005, savran2003, shaari2008, shah2012, shreejith2016, soares2006, sonowal2012, wang2015b, wang2015g, wang2015m, won2007, youssef2012, zhang2005, zhang2006, lazaro2007, abramson1998, atencia2007, boumeridja2005, clemente2016, de2002a, gschwind1996, harmanani2010, mansour2011, saif2006, sousa2014, srinivasulu2012, stepanova2007, varma2002, wakamura2003, schrauwen2008, wang2016l, yi2016, acosta2001, ago2013, aliaga2008, ann2016, bahoura2011, bahoura2011a, benrekia2009, beuchat1998, beuchat1998a, beuchat1999, botros1994, braga2010, canas2006, carvalho2005, chalhoub2006, da2009, denby2003, ferreira2010, ferrer2004, gatet2009a, gomperts2010, gomperts2011, gorgon2006, horita2015, jin2011, johnston2005, khan2006, khodja2010, kim1998, kyoung2006, latino2009, lee2006, leon1999, lin2010, lotrivc2011, nedjah2008, nedjah2009, nedjah2012, nedjah2014, ortigosa2003, ortigosa2003a, ortigosa2006, ortigosa2006a, ozdemir2011, panicker2012, pasero2004, patra2006, perez2014a, qinruo2003, rani2007, rezvani2012, skodzik2013, syiam2003, taright1998, tatikonda2008, vitabile2005, vskoda2011, wolf2001, wolf2001a, yu2006, zhu1999, bahoura2012, bahoura2014, ntoune2012, truong2016a, truong2016b, tisan2010, tisan2015, bezborah2012, cavuslu2012, cawley2010, cawley2011, fe2013, lin2008, low2006, merchant2006, merchant2006a, merchant2008, merchant2010, morgan2009, nambiar2014, upegui2006, aibe2002, aibe2004, bu2004, figueiredo1998, minchin1999, zhou2010, zhu2010, damak2006, hikawa2001, hikawa2002, hikawa2003, hikawa2003a, krid2006, krid2009, lysaght1994, maeda2003, martincigh2005, rossmann1997, schafer1999, vega2006, zhuang2007, halgamuge1994, kim2008, kim2015f,  girau2000, gupta2009, li2015h, lin2009, maeda2005, kim2009, kim2014b, le2010, ben2004, brassai2014, porrmann2002a, tamukoh2010, ahn2013, ahn2015, ahn2015a, ang2011, bako2009, bako2010, bako2013, belhadj2009, bellis2004, bhuiyan2010a, blair2013, caron2011, cassidy2007, cassidy2008, cassidy2011b, chappet2014, cheung2012, cheung2015, cong2013, daffron2016, dahasert2012, de2016a, dean2014, dean2016a, diaz2016, farsa2015, fox2012, garg2011, ghani2006, glackin2005a, glackin2009, glackin2009a, gomar2014, gomar2016, grassia2014, hafliger2001, harkin2008, harkin2009, hashimoto2010, hellmich2004, hellmich2004a, iakymchuk2012, iakymchuk2014, just2010,  koziol2014, li2012, li2013f, makhlooghpour2016, maya2000, molin2015, moore2012, niu2012, nuno2009, nuno2011, nuno2012, pearson2005, pearson2005a, pearson2007, rice2009, rios2016, rodrigues2015, roggen2003, ros2003, rosado2011, rostro2011, rostro2012, sanchez2010, schrauwen2006, shayani2008a, shayani2008b, shayani2009, sheik2016, sofatzis2014, soleimani2012, upegui2005, wang2013, wang2014c, wang2014p, wang2014u, wang2015n, wu2015e, xicotencatl2003, yang2011, yang2011a, zuppicich2009, cao2015, cerezuela2015, bade1994, li2004, nedjah2003, nedjah2003a, nedjah2007, van1993, pearson2006, al2011, al2012, debole2011, kestur2012, prieto2009}.  For many of these implementations, the use of the FPGA is often utilized as a stop-gap solution on the way to a custom chip implementation.  In this case, the programmability of the FPGA is not utilized as part of the neuromorphic implementation; it is simply utilized to program the device as a neuromorphic system that is then evaluated.  However, it is also frequently the case that the FPGA is utilized as the final implementation, and in this case, the programmability of the device can be leveraged to realize radically different network topologies, models, and algorithms.  Because of their relative ubiquity, most researchers have access to at least one FPGA and can work with languages such as VHDL or Verilog (hardware description languages) to implement circuits in FPGAs.  If the goal of developing a neuromorphic system is to achieve speed-up over software simulations, then FPGAs can be a great choice.  However, if the goal is to achieve a small, low-power system, then FPGAs are probably not the correct approach.  Liu and Wang point out several advantages of FPGAs over both digital and analog ASIC implementations, including shorter design and fabrication time, reconfigurability and reusability for different applications, optimization for each problem, and easy interface with a host computer \cite{liu2009a}.  

Full custom or application specific integrated circuit (ASIC) chips have also been very common for neuromorphic implementations \cite{hasan1995, iwata2016, mazumder2016, nanami2016, smith2014a, zhang2013c, knag2016, pedroni2016, salerno1999, lee2014a, christiani2016, distante1990, distante1990a, fornaciari1994, hammerstrom1990, hammerstrom1991, islam2006, kakkar2009, kim1992, kim1993, kumar1996, kung1989, larsson1996, myers1989, pakzad1993, pechanek1994, plaskonos1993, popescu2000, szabo2000, tang1997, tomlinson1990, tomlinson1990a, torbey1992, zhang1991, kuo1993, murtagh1997, ouali1990, ramacher1991, ramacher1991a, ramacher1992, yasunaga1990, yasunaga1991, alla1991, asari1994, blayo1989, blayo1989a, fornaciari1994a, johannet1992,  lehmann1993, masaki1990, pechanek1994, van1989, weinfeld1989, wike1990, yasunaga1993, eickhoff2006, hikawa2001, hikawa2002, kim1995, kim1995a, richert1991, schoenauer1998, tomberg1989, tomberg1990, tomberg1991, maffezzoni1994, suri2015, hung2003, vlontzos1991, wang1991, polepalli2016, polepalli2016a, cambio2003, dlugosz2011a, lehmann1991, marchese2015, mauduit1992, rajah2004, rodriguez2015, ambroise2013, banerjee2015, das2015, dean2016, georgiou2006, hu2014, imam2013, indiveri2007, joubert2012, kim2015c, kulkarni2015, nere2012, nere2013, roclin2013, schoenauer2000, schoenauer2002, seo2015, shen2016, torikai2005, wang2015, zhang2015g, neftci2016, neftci2016a, torralba1995, beichter1993, butler1989, cornu1994, fu1992, gascuel1991, griffin1991, hirai1989, ker1997, lucas1991, mason1992, nakahira1993, ouali1991, uchimura1992, watanabe1993, white1992, yasunaga1989, zhang1997a , chakradhar2010, chen2015c, chen2016c, conti2015, kim2016b, nomura2004, nomura2005, alippi1991a, chung1992, cloutier1994, duranton1986, eguchi1991, kondo1994, madokoro2013, myers1993, oteki1993, saito1998, sato1993, tang1993, theeten1990, wang2006a, wawrzynek1992, yun2002, aihara1996, avellana1996, ayala2002, bermak1999, bermak2002, chang1992, distante1990b, distante1991, duranton1989, faiedh2004, gatet2009, joseph2010, kim2015g, orrey1991, tuazon1993, vincent1991, walker1988,  inigo1990, kim2013a, lewis2003}. IBM's TrueNorth, one of the most popular present-day neuromorphic implementations, is a full custom ASIC design \cite{akopyan2015, arthur2012, cassidy2013a, cassidy2014, imam2012, merolla2011, merolla2014, seo2011}.  The TrueNorth chip is partially asynchronous and partially synchronous, in that some activity does not occur with the clock, but the clock governs the basic time step in the system.  A core in the TrueNorth system contains a 256x256 crossbar configuration that maps incoming spikes to neurons.  The behavior of the system is deterministic, but there is the ability to generate stochastic behavior through pseudo-random source. This stochasticity can be exactly replicated in a software simulation.  

SpiNNaker, another popular neuromorphic implementation, is also a full custom digital, massively parallel system \cite{araujo2014, diehl2014, furber2006a, furber2009, furber2012, furber2013, furber2014, jin2008, jin2010, knight2016, knight2016a, lagorce2015, navaridas2013, painkras2012, painkras2013, plana2011, rast2009, rast2010, sharp2011, sharp2011a, sharp2012, sharp2013, stromatias2013, stromatias2015}.  SpiNNaker is composed of many small integer cores and a custom interconnect communication scheme which is optimized for the communication behavior of a spike-based network architecture.  That is, the communication fabric is meant to handle a large number of very small messages (spikes).  The processing unit itself is very flexible and not custom for neuromorphic, but the configuration of each SpiNNaker chip includes instruction and data  memory in order to minimize access time for frequently used data.  Like TrueNorth, SpiNNaker supports the cascading of chips to form larger systems.  

TrueNorth and SpiNNaker provide good examples of the extremes one can take with digital hardware implementations.  TrueNorth has chosen a fixed spiking neural network model with leaky integrate-and-fire neurons and limited programmable connectivity, and there is no on-chip learning.  It is highly optimized for the chosen model and topology of the network.  SpiNNaker, on the other hand, is extremely flexible in its chosen neuron model, synapse model, and learning algorithm.  All of those features and the topology of the network are extremely flexible. However, this flexibility comes at a cost in terms of energy efficiency.  As reported by Furber in \cite{furber2016}, TrueNorth consumes 25 pJ per connection, whereas SpiNNaker consumes 10 nJ per connection.

\subsubsection{Analog}

Similar to the breakdown of digital systems, we separate analog systems into programmable and custom chip implementations.  As there are FPGAs for digital systems, there are also field programmable analog arrays (FPAAs).  For many of the same reasons that FPGAs have been utilized for digital neuromorphic implementations, FPAAs have also been utilized \cite{dahasert2012, dong2006, marr2015, mcginley2008, nease2012, rocke2005, rocke2007, rocke2008, shapero2013, yao2013, zhao2007}. There have also been custom FPAAs specifically developed for neuromorphic systems, including the field programmable neural array (FPNA) \cite{farquhar2006} and the NeuroFPAA \cite{liu2009b}.  These circuits contain programmable components for neurons, synapses, and other components, rather than being more general FPAAs for general analog circuit design. 


It has been pointed out that custom analog integrated circuits and neuromorphic systems have several characteristics that make them well suited for one another.  In particular, factors such as conservation of charge, amplification, thresholding and integration are all characteristics that are present in both analog circuitry and biological systems \cite{mead1990}.  In fact, the original term neuromorphic was used to refer to analog designs.  Moreover, taking inspiration from biological neural systems and how they operate, neuromorphic based implementations have the potential to overcome some of the issues associated with analog circuits that have prevented them from being widely accepted.  Some of these issues are dealing with global asynchrony and noisy, unreliable components \cite{indiveri2002b}.  For both of these cases, systems such as spiking neural networks are natural applications for analog circuitry because they can operate asynchronously and can deal with noise and unreliability.  

 One of the common approaches for analog neuromorphic systems is to utilize circuitry that operates in subthreshold mode, typically for power efficiency purposes \cite{alvado2001, andreou1989, andreou1990, andreou1990a, andreou1994, azghadi2013c, badoni2006, bartolozzi2006a, bartolozzi2008, boahen1989, boahen1989a, camilleri2010, cauwenberghs1996, cauwenberghs1996a, chasta2012, cheely2003, chen2012b, chien2015, choi1993, coue1996, cymbalyuk2000, el2003, elias1994a, gallagher2005, georgiou1999, ghaderi2013, giulioni2011, holleman2015a, huayaney2016, indiveri1994, indiveri1999a, lan1995, lee2007, liu2000, lu2015, maher1989, maliuk2010, maliuk2014a, mandloi2014, maliuk2014a, matolin2004a, meng2011, morie1994, morie1999, morris1998a, nakada2005, pan2012, papadimitriou2014, pinjare2009, schreiter2002, sitte2007, song1993, valle1994, yu2010, yu2010b}.   In fact, the original neuromorphic definition by Carver Mead referred to analog circuits that operated in subthreshold mode \cite{mead1990}.  There are a large variety of other neuromorphic analog implementations \cite{howard1987, kaulmann2005, linares1993, maundy1991a, mccarley1995, rossetto1989, saeki2011, alvado2004, basu2010a, chen2010,  douence1999, elias1992, elias1992a, grassia2011, huayaney2016a, jung2001, kanazawa2003, laflaquiere1997, laflaquiere1997a, lewis2000, linares1994, ma2013, mayr2014, millner2012, park2016, passetti2013, pinto2000, rachmuth2003, rachmuth2008, renaud2004, rovere2014, ryckebusch1989, saighi2005, saighi2005a, saighi2006, saighi2010a, saighi2011, santurkar2015, schultz1995, serrano2008, sharma2016, shi2007, simoni2006, still2006, tomas2006, vanschaik2004, vittoz1990, wang2013b, yu2009, yu2009a, yu2011, yu2012c, zhou2013, jayakumar1992, madani1991, pujol1994, schneider1991b, schneider1993, cruz1991, cruz1998, dalla1993, halonen1991, harrer1992, kinget1994, kinget1995, rodriguez1993, yang1990, martins1998, holleman2015, abutalebi1998, akers1989, aksin2009, al2001, alhalabi1995, almeida1996, alspector1993, amaral2004, arima1992, bayraktarouglu1999, berg1996, bibyk1989, bibyk1990, bo1996, bo1997, bo1999, bollano1997, borgstrom1990, botha1992, bridges2005, cairns1994, calayir2015, carvajal2011, chang1996,cho1996, cho1998, cho1999, choi1991, choi1992, choi1993a, choi1993b, choi1996, coggins1994, coggins1995, coggins1995a, diotalevi2000, docheva2007, dolenko1993, dolenko1993a, dolenko1995, donald1993, duong1992, eberhardt1989, eberhardt1992, el1997, fakhraie1995, feltham1991, fisher1990, foruzandeh1999, foruzandeh1999a, furman1988, gatet2006, gatet2007, gatet2008, gatet2008a, gatet2008b, gatet2009, heruseto2009, hirotsu1993, hohmann2002, houselander1988, jabri1992, kakkar2009, lansner1992, leong1992, lindblad1995, liu1999, liu1999a, lont1992, lont1993, lu2001, lu2001a, lu2001b, lu2002, lu2002a, lu2002b, lu2003, maeda1993, maeda1995, maliuk2010a, maliuk2012, maliuk2014, masa1994, masmoudi1999, mestari2004, michel2004, milev2003, modi2006, montalvo1994, montalvo1997, montalvo1997a, morns1999, mundie1994, murray1992a, nosratinia1992a, oh1993, oh1994, richter2015, salam1990, satyanarayana1989, shimabukuro1989, soelberg1994, song1994, sun2002, tam1990, tam1992, tawel1993, thakur2015a, thakur2016, tombs1993, tsividis1987, valle1996, van1992, walker1989, wang1993, wawryn2001, wawryn2001a, weller1990, withagen1994, wolpert1992, yildirim1996, yildiz2007, castro1993, damle1997, etienne1994, holler1989,  mueller1989, mueller1989a, card1994, cauwenberghs1991, schneider1991, schneider1991a, schneider1991c, shima1992, kamio1997, khachab1989, lee1990a, linares1992, linares1992a, paulos1988, verleysen1989, yang1994, woodburn2000, churcher1993, woodburn1994, dogaru1996, verleysen1994, yildirim1996, abdelbaki2000, boddhu2006, brownlow1991, cauwenberghs1994, fisher1991, gallagher2000, gallagher2001, gallagher2008, hasler1991, kothapalli2005, lansner1993, salam1989, salam1989a, salam1991, schemmel2001, thakoor1991, dlugosz2010, mann1989, maundy1991, sheu1992, annunziato1998, bartolozzi2004,  binas2016, bofill2001, bofill2003, bofill2004, camilleri2007, carota2012, chicca2001, chicca2003, chicca2003a, dungen2005, fieres2008, fusi1999, fusi2000a, garg2011, ghaderi2015, giulioni2008, giulioni2008a, giulioni2009, giulioni2015, giulioni2016, glover1998, glover1999, glover2002, goldberg2001, goldberg2001a, gordon2002, grassia2011a, hafliger1997, hafliger1999a, hafliger2007, hajtavs2000, hsieh2013, hsieh2013a, huayaney2011, huo2005, indiveri2000, indiveri2000a, indiveri2001b, indiveri2002a, indiveri2007, joubert2011, joubert2012, kanazawa2004, koickal2007, koickal2009, lazzaro1992, liu2006, mahowald1997, mill2011, mitra2006, nease2013, neftci2013, oster2004, oster2005, oster2008, perez2014, pfeil2013a, pfeil2016, querlioz2013a, renaud2007, riis2004, rodrigues2014, saighi2010, schemmel2004, schemmel2007, schemmel2008, schemmel2012, smith1998, sun2011, tanaka2009, tovar2006, tovar2008, utagawa2007, vanschaik2001, vogelstein2002, wang2006, wang2009, wang2010, wijekoon2012a, yang2004, zhang2010, zhao2016b, gatt2000, van1994, al1999, andreou1994a, badoni1994, bo2000, brownlow1990, buhlmeier1996, cichocki1992, conti1996, daniell1989, dlugosz2011, ghosh1994, hamilton1989, hamilton2011, hasler1990, heittmann2002a, ho1994, houselander1989, jackson1994, mirhassani2008, moon1991, morris1998, mosin2015, mosin2016, mueller1992, murray1989, nakada2005a, nakada2006, onorato1994, pelayo1990, pelayo1995, reyneri1994, rodriguez1988, rodriguez1990, roy1999, salam1989b, saxena1994, schwartz1989, serrano1996, shibata1992, torralba1999, tsividis1987a, wilson1996, wojtyna2006, wojtyna2008, yanai1990, andreou1990a, andreou1995, arreguit1994, cosp1999, cosp2003, graf1995, horiuchi1999, indiveri1997, indiveri1999, liu2001, markan2007, chen2007, chen2010a, matolin2004, ota1999, xiong2010, nawrocki2011}.

\subsubsection{Mixed Analog/Digital}

Mixed analog/digital systems are also very common for neuromorphic systems \cite{al2000, arthur2004, arthur2007, asai2004a, barkan1990, bor1996, chen1996, current1990, hahnloser2000, hajtas2004, han2005, han2006, han2006a, han2007, han2016b, heittmann2002, huo2012, jackel1987, johnson1995, kang2015, kaplan2009, korekado2003, lee1990b, lee1991, liu2002a, liu2004, liu2010a, lu2009, maeda1999, mann1988, maundy1990, maundy1991b, mayr2014a, mayr2016a, mcdaid2008, meier2015, merolla2003, merolla2006, moradi2011, moradi2014, morie2000, murray1988, noack2015, ota1996, ozalevli2005, palma2013, pan2003, pan2003a, patel2000a,  petre2016, pfeil2013, qiao2015, renaud2010, ruckert1988, ruckert1991, sackinger1991, sackinger1992, sato1992, schemmel2006, schemmel2010, schmuker2014, sheik2012a, shi2015a, shin1993, sinha2010, sinha2011, stromatias2015a, tapson2012, tarassenko1991, tenore2003, tenore2004, tenore2005, tenore2006, tomberg1989a, tomberg1990, tomberg1992, viredaz1994, vogelstein2007a, vogelstein2008, waller1991, wang2006b, wang2011a, wang2014b, wang2015, wang2016g, watanabe1997, wijekoon2012, wilamowski1996, wu1996, xu2015, yazdi1993, yentis1996, zahn1996, zhang2005a}.  Because of its natural similarity to biological systems, analog circuitry has been commonly utilized in mixed analog/digital neuromorphic systems to implement the processing components of neurons and synapses.   However, there are several issues with analog systems that can be overcome by utilizing digital components, including unreliability.  

In some neuromorphic systems, it has been the case that synapse weight values or some component of the memory of the system are stored using digital components, which can be less noisy and more reliable than analog-based memory components \cite{abinaya2015, aibara1991, azghadi2013a, azghadi2015, bruderle2010, camboni2001, cardarilli1994, cardarilli1995, corradi2015, corso1990, corso1992, del1990, del1990a, del2008, djahanshahi1996, djahanshahi1996a, djahanshahi1996a, djahanshahi1997, djahanshahi1997a, elias1993, elias1994, elias1995, elias1995a, fang1990, fang1992, franca1993, friedmann2013, horio2003, hylander1993, koosh2001, koosh2002, lee1992, lehmann1996, lenero2008, lu2007, masa1994a, minkovich2012, mirhassani2003, mirhassani2004, mirhassani2005, moopenn1990, murray1987, neugebauer1991, nosratinia1992, passos1993, petrovici2014, raffel1989, satyanarayana1992, schmid1999a, sibai1997, tank1986, van1990, yang1999, zaman1994, zatorre2006}.  For example, synaptic weights are frequently stored in digital memory for analog neuromorphic systems.  Other neuromorphic platforms are primarily analog, but utilize digital communication, either within the chip itself, to and from the chip, or between neuromorphic chips \cite{aamir2016, afifi2011, bruderle2010, charles2008, chicca2004, chicca2007a, choi2005, choudhary2012, corneil2012, corneil2012a, corradi2014, del1998, deyong1992, deyong1992a, deyong1993, deyong1994, donati2014, donati2016, douglas1994, ehrlich2007, fakhraie2004, fieres2004, graf1987a, hammerstrom2009, higgins1999, higgins2000, hock2013, hollis1990, indiveri2001a, indiveri2004, indiveri2006, indiveri2011b, indiveri2015c, mitra2007a, mitra2009, mitra2009a, mostafa2014, mumford1992, murray1991, neftci2007, neftci2010, neftci2010a, neftci2011, neftci2011a, neftci2012, neftci2012a, sanchez2013, schemmel2004a, shi2003a, vogelstein2004, vogelstein2004a, vogelstein2007, wang2007, yang2012a, yu2012b, arima1991a, bartolozzi2007a, basu2013, boi2016, boser1991, brink2013, folowosele2008, graf1988, graf1990, hansen1989, mitra2007, zou2006, zou2006a}.  Communication within and between neuromorphic chips is usually in the form of digital spikes for these implementations. Using digital components for programmability or learning mechanisms has also been common in mixed analog/digital systems \cite{alvado2003, arena2006, arthur2006, boi2016, chen2003, chen2003, chicca2006, douence2001, erkmen2013, jackel1990, jackel1990a, lee1990, lenero2008a, linares2003, arima1991, beerhold1990, cruz2012, hussain2012, schmid1999}.

Two major projects within the mixed analog/digital family are Neurogrid and BrainScaleS.  Neurogrid is a primarily analog chip that is probably closest in spirit to the original definition of neuromorphic as coined by Mead \cite{benjamin2014, boahen2006, menon2014, merolla2014a}.  Both of these implementation fall within the mixed analog/digital family because of their digital communication framework.  BrainScaleS is a wafer-scale implementation that has analog components \cite{hock2013, meier2015, millner2012, pfeil2012, pfeil2013, pfeil2016, schemmel2012}.  Neurogrid operates in subthreshold mode, and BrainScaleS operates in superthreshold mode.  The developers of BrainScaleS chose superthreshold mode because it allows BrainScaleS chips to operate at a much higher rate than is possible with Neurogrid, achieving a 10,000x speed-up \cite{furber2016}.


\subsection{Device-Level Components}
\label{sec:devicelevel}

\begin{figure}[!t]
\centering
\includegraphics[width=0.5\textwidth]{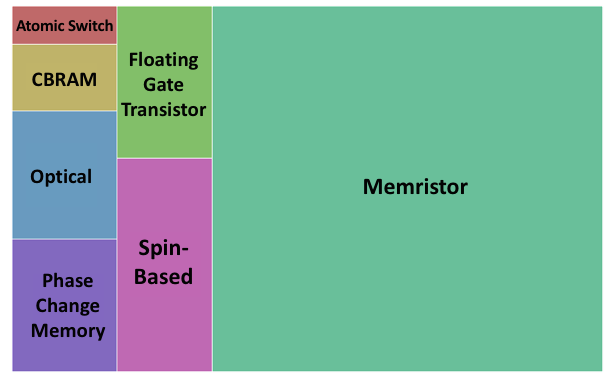}
\caption{Device-level components and their relative popularity in neuromorphic systems.  The size of the boxes corresponds to the number of works referenced that have included those components.}
\label{fig:device_components_treemap}
\end{figure}

In this section, we cover some of the non-standard device level or circuit level components that are being utilized in neuromorphic systems.  These include a variety of components that have traditionally been used as memory technologies, but they also include elements such as  optical components.  Figure \ref{fig:device_components_treemap} gives an overview of the device-level components and also shows their relative popularity in the neuromorphic literature.

\subsubsection{Memristors}
\label{sec:memristors}
Perhaps the most ubiquitous device-level component in neuromorphic systems is the ``memory resistor'' or the memristor.  Memristors were a theoretical circuit element proposed by Leon Chua is 1971 \cite{chua1971memristor} and ``found'' by researchers at HP in 2008 \cite{strukov2008missing}. The key characteristic of memristive devices is that the resistance value of the memristor is dependent upon its historical activity.  One of the major reasons that memristors have become popular in neuromorphic computing is their relationship to synapses; namely, circuits that incorporate memristors can exhibit STDP-like behavior that is very similar to what occurs in biological synapses.  In fact, it has been proposed that biological STDP can be explained by memristance \cite{linares2009a}.  Memristors can be and have been made from a large variety of materials, some of which will be discussed in Section \ref{sec:materials}, and these different materials can exhibit radically different characteristics.  Another reason for utilizing memristors in neuromorphic systems is their potential for building energy efficient circuitry, and this has been studied extensively, with several works focused entirely on evaluating energy consumption of memristive circuitry in neuromorphic systems \cite{deng2016, hsieh2016, liu2013c, rajendran2013, taha2013, taha2014a, tang2014, wang2015i}.  It has also been observed that neuromorphic implementations are a good fit for memristors because the inherent fault tolerance of neural network models can help mitigate effects caused by memristor device variation \cite{chen2014b, gao2014a, querlioz2011, querlioz2013, zhu2011}.  


A common use of memristors in neuromorphic implementations is as part of or the entire synapse implementation (depending on the type of network) \cite{pershin2010, wang2015j, indiveri2013, mandal2014, wang2012a, zhu2017, bojnordi2016, rafique2016, sheri2015, suri2015a, ascoli2016, corinto2014, duan2014a, hu2012, shi2015b, garbin2014, garbin2015, garbin2015a, merkel2014, merkel2014b, suri2015b, burger2014, dong2014, emelyanov2015, emelyanov2016, manem2015, merkel2014c, such2015, ibrayev2016, duan2015, guo2015b, hu2015, liu2012, liu2013b, liu2015d, jackson2015, jackson2015a, jackson2016, sharma2015a, serb2016, kudithipudi2015, cai2014a, chang2011a, chua2013, corinto2013, corinto2013a, gaba2013, hu2016a, naous2016, kang2015a, lorenzi2015, wu2013a}.  Sometimes the memristor is simply used as a synaptic weight storage element.  In other cases, because of their plasticity-like properties, memristors have been used to implement synaptic systems that include Hebbian learning in general \cite{aggarwal2012, ascoli2014, cantley2011, cantley2012, kubendran2012, wang2014q, wen2013b, ziegler2013} or STDP in particular \cite{acciarito2016, afifi2009, babacan2017, berdan2016, bill2014, cai2012, cai2015, cantley2011a, chan2012, chen2014a, covi2015, covi2016a, cruz2013, dai2014, desalvo2015, du2015a, ebong2010, ebong2011, ebong2012, gale2012, he2014, hu2013c, hu2016b, jo2010, kaneko2013, kaneko2014, kavehei2011, lecerf2013, li2015c, li2015f, mandal2013, moon2014, mostafa2015, pantazi2016, payvand2014, payvand2015, perez2010, prezioso2016a, seo2011a, serrano2012, serrano2014, singha2014, snider2008, subramaniam2013, wang2014r, werner2016, werner2016a, wozniak2016, zhang2014, zheng2015}.  Perhaps the most common use of a memristor in neuromorphic systems is to build a memristor crossbar to represent the synapses in the network \cite{adam2017, afifi2009a, agarwal2015, azghadi2016, bavandpour2015, bennett2016, chabi2011, chabi2014, chabi2014a, chabi2015, chabi2015a, chen2013b, chen2016a, chen2016b, chi2016, choi2009, chu2015, hasan2014, hasan2015, hu2012a, hu2014a, hu2016, jiang2016, kataeva2015, kim2011, kim2012a, kim2015a, li2014, li2014d, li2015b, li2015d, li2015e, li2016c, linares2009, liu2013, liu2014, liu2014a, liu2015, liu2015a, liu2015c, liu2016a, liu2016c, long2016, merkel2015a, merrikh2014, mostafa2016, nair2015, naous2016a, park2012, park2013, park2013a, park2015, prezioso2015, prezioso2015a, prezioso2016, prezioso2016b, prezioso2016d, qiu2014, qiu2015, ren2013, rothganger2015, sengupta2015, shafiee2016, sheri2014, starzyk2013, starzyk2014, taha2014, tarkov2015, tarkov2015a, truong2014, truong2014a, truong2016, truong2016a, truong2016b, wang2014f, wang2014i, wang2014j, wang2016g, wang2016i, wheeler2011, wheeler2014, wu2015, wu2015d, wu2015g, xia2016a, xie2016, xu2015a, yakopcic2013, yakopcic2014, yakopcic2014a, yakopcic2014b, yakopcic2015, yakopcic2015a, yakopcic2015c, yakopcic2015d, yakopcic2015e, yakopcic2016, yan2016, yang2013, yao2015, yogendra2015, yu2012a, yu2015, zamanidoost2015}.  Early physical implementations of memristors have been in the crossbar configuration. Crossbar realizations are popular in the literature mainly due to their density advantage by also because physical crossbars have been fabricated and shown to perform well. Because a single memristor cannot represent positive and negative weight values for a synapse (which may be required over the course of training for that synapse), multi-memristor synapses have been proposed, including memristor-bridge synapses, which can realize positive, negative and zero weight values \cite{adhikari2012, adhikari2014, adhikari2015, kim2012b, kim2014, sah2012, sah2012a, sah2013, tarkov2016, tarkov2016a, wang2015p}.  Memristor-based synapse implementations that include forgetting effects have also been studied \cite{chen2013a, zhang2017}.  Because of their relative ubiquity in neuromorphic systems, a set of training algorithms have been developed specifically with characteristics of memristive systems in mind, such as dealing with non-ideal device characteristics \cite{chen2015a, chen2015b, danilin2015, danilin2015a, deng2015, ebong2010, ebong2012, howard2011, howard2012, howard2014, howard2014a, howard2015, liu2013a, merkel2013, querlioz2011a, querlioz2012, rose2011a, rosenthal2016, shahsavari2016, soltiz2012, soltiz2013, soudry2015, woods2014, woods2015, wu2016}.

Memristors have also been utilized in neuron implementations \cite{al2015, al2015a, babacan2016, demin2015, mehonic2016, pantazi2016, shamsi2015, wang2013d, galushkin2014}.  For example, memristive circuits have been used to generate complex spiking behavior \cite{feali2016a, gale2014, gale2014a}.  In another case, a memristor has been utilized to add stochasticity to an implemented neuron model \cite{suri2015a}.  Memristors have also been used to implement Hodgkin-Huxley axons in hardware as part of a neuron implementation \cite{feali2016, pickett2013}. 


It is worth noting that there are a variety of issues associated with using memristors for neuromorphic implementations.  These include issues with memristor behavior that can seriously affect the performance of STDP \cite{bayat2013, bayat2015, serb2014}, sneak paths \cite{gi2015}, and geometry variations \cite{pino2012}. It is also worth noting that a fair amount of theory about memristive neural networks has been established, including stabilization \cite{bao2013, cai2016, chandrasekar2016, chen2014c, chen2014d, chen2014e, chen2015, duan2016, guo2013, hu2010, jiang2015b, li2014b, li2014e, li2014f, li2016, mathiyalagan2015a, meng2016, qi2014a, rakkiyappan2014b, rakkiyappan2014c, vasilkoski2011, wang2014a, wang2014l, wang2014m, wang2014o, wang2015a, wang2015e, wang2015r, wang2016d, wang2016e, wen2012, wen2012a, wen2013d, wen2014, wen2015, wu2011, wu2012, wu2012a, wu2012d, wu2013b, wu2014a, wu2014d, wu2014e, wu2014f, wu2014g, wu2014h, wu2015a, wu2015c, wu2017, xin2016, yang2014a, yang2015, yang2016b, zhang2012, zhang2013, zhang2014b, zhang2015a, zhang2016a, zhong2016}, synchronization \cite{abdurahman2015, abdurahman2015a, anbuvithya2015, ascoli2014a, ascoli2015, bao2015, bao2015a, bao2015b, bao2015c, bao2016, chandrasekar2014, chen2014c, chen2015, corinto2011, ding2015, ding2016, ding2016a, guo2014c, guo2015, guo2016, han2016, heittmann2011, jiang2015, jiang2015a, li2015, liu2015b, liu2016, mathiyalagan2015, mathiyalagan2016, sakthivel2016, shi2014, song2015, velmurugan2015, velmurugan2016, volos2011, volos2015, wan2015, wang2014, wang2014h, wang2014m, wang2014t, wang2015c, wang2015f, wang2015h, wang2016, wang2016f, wen2013, wen2014a, wu2011a, wu2012b, wu2013, wu2013c, wu2014j, wu2015b, wu2015f, yan2015, yang2014, yang2014a, yang2015d, yang2015e, yang2016a, yang2016b, zhang2013a, zhang2013b, zhang2014a, zhang2015b, zhang2015c, zhang2015f, zhang2016a, zhao2015b}, and passivity \cite{ali2015, anbuvithya2016, duan2014, guo2014a, li2015a, li2016a, meng2015, rakkiyappan2014, rakkiyappan2014a, rakkiyappan2015, velmurugan2014, wang2014n, wen2013a, wu2014b, wu2014c, xiao2015, zhang2015} criteria for memristive neural networks.  However, these works are typically done with respect to ideal memristor models and may not be realistic in fabricated systems.

\subsubsection{CBRAM and Atomic Switches}

Conductive-bridging RAM (CBRAM) has also been utilized in neuromorphic systems.  Similar to resistive RAM (ReRAM), which is implemented using metal-oxide based memristors or memristive materials, CBRAM is a non-volatile memory technology.  CBRAM has been used to implement synapses \cite{burr2017, bichler2013, clermidy2014, desalvo2015, desalvo2015a, gamrat2015, mahalanabis2016, roclin2014, suri2012,  suri2013, suri2015b, yu2010a} and neurons \cite{jang2016, palma2013}. CBRAM differs from resistive RAM in that it utilizes electrochemical properties to form and dissolve connections.  CBRAM is fast, nanoscale, and has very low power consumption \cite{burr2017}.  Similarly, atomic switches, which are nano-devices related to resistive memory or memristors, control the diffusion of metal ions to create and destroy an atomic bridge between two electrodes \cite{aono2010atomic}.  Atomic switches have been fabricated for neuromorphic systems.   Atomic switches are typically utilized to implement synapses and have been shown to implement synaptic plasticity in a similar way to approaches with memristors \cite{avizienis2012, hasegawa2010, nayak2012, sillin2013, stieg2014, tsuruoka2012, tsuruoka2014, yang2013a}.

\subsubsection{Phase Change Memory}

Phase change memory elements have been utilized in neuromorphic systems, usually to achieve high density.  Phase change memory elements have commonly been used to realize synapses that can exhibit STDP.  Phase change memory elements are usually utilized for synapse implementations \cite{ambrogio2016a, eryilmaz2014, garbin2013, jackson2013, kang2015b, kim2015d, shelby2015, suri2011, suri2012a, suri2012b, suri2013a, suri2013b, wang2015q, zhong2015, eryilmaz2013, bichler2012, kuzum2011, kuzum2011a, kuzum2012, lee2014, skelton2015} or synapse weight storage \cite{burr2014, burr2015, burr2015a, sidler2016}, but they have also been used to implement both neurons and synapses \cite{tuma2016, tuma2016a, wright2013}. 

\subsubsection{Spin Devices}

One of the proposed beyond-CMOS technologies for neuromorphic computing is spintronics (i.e., magnetic devices).  Spintronic devices and components have been considered for neuromorphic implementation because they allow for a variety of tunable functionalities, are compatible with CMOS, and can be implemented at nanoscale for high density.  The types of spintronic devices utilized in neuromorphic systems include spin-transfer torque devices,  spin-wave devices, and magnetic domain walls \cite{roy2014b,  locatelli2014 , grollier2016, roy2015a, sengupta2016}.  Spintronic devices have been used to implement neurons \cite{liyanagedera2016, nakada2016, yogendra2015, fan2015, sengupta2015, sengupta2015a, sengupta2016e, sengupta2016h, sharad2012b}, synapses that usually incorporate a form of on-line learning such as STDP \cite{sengupta2015b, sengupta2016c, adachi2015, sengupta2016f, lequeux2016, suh2015, vincent2014, zhang2015e}, and full networks or network modules \cite{roska2012, sharad2012, sengupta2016a, ramasubramanian2014, sengupta2016b, sharad2013, sharad2013a, locatelli2015, sengupta2016d, sengupta2016g, sengupta2016j, zhang2016d, sengupta2016i, sharad2012a, sharad2012c, zeng2016, zhang2016b, diep2014}.
\subsubsection{Floating Gate Transistors} 

Floating-gate transistors, commonly used in digital storage elements such as flash memory \cite{hasler2001floating-gate}, have  been utilized frequently in neuromorphic systems.  As Aunet and Hartmann note, floating-gate transistors can be utilized as analog amplifiers, and can be used in analog, digital, or mixed-signal circuits for neuromorphic implementation \cite{aunet2003}.  The most frequent uses for floating-gate transistors in neuromorphic systems have been either as analog memory cells for synaptic weight and/or parameter storage \cite{durfee1992, fabbrizio1996, fowler1994, fujita1993, hafliger1999, harrison2001, rahimi2002, sin1992, morie2003} or as a synapse implementation that usually includes a learning mechanism such as STDP \cite{brink2013a, diorio1997, diorio1997a, diorio1998, gupta2014, hasler1998, hasler2001, hindo2014, kosaka1995, lee1991a, liu2008, markan2013, pankaala2009, ramakrishnan2011, riggert2014, shibata1995}.   However, floating gate transistors have also been used to implement a linear threshold element that could be utilized for neurons \cite{aunet2003}, a full neuron implementation \cite{wong2007}, dendrite models \cite{brink2008}, and to estimate firing rates of silicon neurons \cite{nease2015}. 

\subsubsection{Optical}
Optical implementations and implementations that include optical or photonic components are popular for neuromorphic implementations \cite{brunner2016, coomans2011, frye1991, maier1999, neiberg1994, paquot2012, prucnal2011, de2016, shen2016a, fok2011}.  In the early days of neuromorphic computing, optical implementations were considered because they are inherently parallel, but it was also noted that the implementation of storage can be difficult in optical systems \cite{collins1989}, so their implementations became less popular for several decades.  In more recent years, optical implementations and photonic platforms have reemerged because of their potential for ultrafast operation, relatively moderate complexity and programmability \cite{shastri2016, fok2016}. Over the course of development of neuromorphic systems, optical and/or photonic components have been utilized to build different components within neuromorphic implementations.  Optical neuromorphic implementations include optical or opto-electronic synapse implementations in early neuromorphic implementations \cite{cauwenberghs1990, rietman1989} and more recent optical synapses, including using novel materials \cite{gholipour2015, livingston1991, maier2016, qin2016, ren2015}.  There have been several proposed optical or photonic neuron implementations in recent years \cite{fok2012, hurtado2012, kravtsov2011, nahmias2013, nahmias2016, romeira2016, shainline2016}.  

\subsection{Materials for Neuromorphic Systems}
\label{sec:materials}

One of the key areas of development in neuromorphic computing in recent years have been in the fabrication and characterization of materials for neuromorphic systems.  Though we are primarily focused on the computing and system components of neuromorphic computing, we also want to emphasize the variety of new materials and nano-scale devices being fabricated and characterized for neuromorphic systems by the materials science community.  

Atomic switches and CBRAM are two of the common nano-scale devices that have been fabricated with different materials that can produce different behaviors.  A review of different types of atomic switches for neuromorphic systems is given in \cite{tsuruoka2014}, but common materials for atomic switches are \ce{Ag_2S} \cite{avizienis2012, hasegawa2010, stieg2014}, \ce{Cu_2S}  \cite{nayak2012}, \ce{Ta_2O_5} \cite{tsuruoka2012}, and \ce{WO_{3-x}} \cite{yang2013a}.   Different materials for atomic switches can exhibit different switching behavior under different conditions.  As such, the selection of the appropriate material can govern how the atomic switch will behave and will likely be application-dependent.  CBRAM has been implemented using \ce{GeS_{2}/Ag} \cite{desalvo2015, suri2012, suri2013, suri2015b, clermidy2014, palma2013, roclin2014}, \ce{HfO_{2}/GeS_2} \cite{desalvo2015a}, \ce{Cu/Ti/Al_2O_3} \cite{jang2016}, \ce{Ag/Ge_{0.3}Se_{0.7}} \cite{mahalanabis2014, mahalanabis2016, yu2010a}, \ce{Ag_2S} \cite{la2015, la2016, labarbera2015} and \ce{Cu/SiO_2} \cite{yu2010a}. Similar to atomic switches, the switching behavior of CBRAM devices is also dependent upon the material selected; the stability and reliability of the device is also dependent upon the material chosen.  

There are a large variety of implementations of memristors.  Perhaps the most popular memristor implementations are based on transition metal-oxides (TMOs).  For metal-oxide memristors, a large variety of different materials are used, including \ce{HfO_x} \cite{covi2016, gao2014, gao2015a, jha2014, matveyev2016, yu2011a, matveyev2015, woo2016a, jia2014}, \ce{TiO_x} \cite{demin2016, dongale2016, hu2014b, hu2014d, park2016a, okelly2016}, \ce{WO_x} \cite{chang2012, du2015, tan2015a, shi2016, thakoor1990}, \ce{SiO_x} \cite{chang2016a, guo2013a}, \ce{TaO_x/TiO_x} \cite{gao2015, wang2015k}, \ce{NiO_x} \cite{hu2013, hu2013b, hu2014c}, \ce{TaO_x} \cite{yang2015c, wang2016c, thomas2015}, \ce{FeO_x} \cite{wang2016b}, \ce{AlO_x} \cite{wu2012c, sarkar2015}, \ce{TaO_x/TiO_x} \cite{gao2015, wang2015k}, \ce{HfO_x/ZnO_x} \cite{wang2017b}, and PCMO \cite{moon2015, jang2014, jang2015, lee2015a, moon2016, moon2016a} . Different metal oxide memristor types can produce different numbers and types of resistance states, which govern the weight values that can be ``stored" on the memristor.  They also have different endurance, stability, and reliability characteristics.  

A variety of other materials for memristors have also been proposed.  For example, spin-based magnetic tunnel junction memristors based on \ce{MgO} have been proposed for implementations of both neurons and synapses \cite{krzysteczko2012}, though it has been noted that they have a limited range of resistance levels that make them less applicable to store synaptic weights \cite{ thomas2015}.  Chalcogenide memristors \cite{li2013, li2014a, tranchant2015} have also been used to implement synapses; one of the reasons given for utilizing chalcogenide-based memristors is ultra-fast switching speeds, which allow for processes like STDP to take place at nanosecond scale \cite{li2013}. Polymer-based memristors have been utilized because of their low cost and tunable performance \cite{chen2014, demin2016, juarez2016, li2013b, luo2015, luo2016, nawrocki2014, xiao2016, yang2016, zhang2016}.  Organic memristors (which include organic polymers) have also been proposed \cite{bennett2015, cabaret2014, chang2016, demin2016, erokhin2010, erokhin2013, erokhina2015, kim2016c, kong2016, lin2016, liu2015j, nawrocki2010, wang2016n}.  



Ferroelectric materials have been considered for building analog memory for synaptic weights \cite{ishiwara1993, yoon1999, yoon1999a, yoon1999b, yoon2000, yoon2000a}, and synaptic devices \cite{kim2016a, nishitani2012, nishitani2013, yoon2016}, including those based on ferroelectric memristors \cite{nishitani2014, nishitani2015, wang2014g}.   They have primarily been investigated as three-terminal synaptic devices (as opposed other implementations that may be two-terminal).  Three-terminal synaptic devices can realize learning processes such as STDP in the device itself \cite{kim2016a, nishitani2015}, rather than requiring additional circuitry to implement STDP.  

Graphene has more recently been incorporated in neuromorphic systems in order to achieve more compact circuits.  It has been utilized for both transistors \cite{wan2016c, wan2016e, yang2016e} and resistors \cite{darwish2016} for neuromorphic implementations and in full synapse implementations \cite{ tian2015, wang2017c}.

Another material considered for some neuromorphic implementations is the carbon nanotube.  Carbon nanotubes have been proposed for use in a variety of neuromorphic components, including dendrites on neurons \cite{hsu2010, hsu2014, hsu2014a, joshi2009, parker2008}, synapses \cite{barzegarjalali2015, barzegarjalali2016, barzegarjalali2016a,  barzegarjalali2016b, friesz2007, gacem2013, joshi2009a, joshi2011, kim2013, kim2015b, kim2015e, liao2011, mahvash2011, shen2013, shen2015, yin2016, zhao2010, feng2016}, and spiking neurons \cite{chen2012, joshi2010a, mahvash2013, najari2016}.  The reasons that carbon nanotubes have been utilized are that they can produce both the scale of neuromorphic systems (number of neurons and synapses) and density (in terms of synapses) that may be required for emulating or simulating biological neural systems.  They have also been used to interact with living tissue, indicating that carbon-nanotube based systems may be useful in prosthetic applications of neuromorphic systems \cite{joshi2011}. 

A variety of synaptic transistors have also been fabricated for neuromorphic implementations, including silicon-based synaptic transistors \cite{kim2016, kim2016d} and oxide-based synaptic transistors \cite{shao2016, shi2013,  wan2013, wan2014, wan2014a, wan2015a, wan2016, wan2016a, wang2016o, zhou2015a, zhu2015, zhu2016a, zhou2013a}. Organic electrochemical transistors \cite{gkoupidenis2015, gkoupidenis2015a, qian2016, wan2016b, wood2012, xu2016 } and organic nanoparticle transistors \cite{alibart2010, alibart2012, bichler2010, kwon2013} have also been utilized to build neuromorphic components such as synapses.  Similar to organic memristors, organic transistors are being pursued because of their low-cost processing and flexibility.  Moreover, they are natural for implementations of brain-machine interfaces or any kind of chemical or biological sensor \cite{gkoupidenis2015}.  Interestingly, groups are pursuing the development of transistors within polymer based membranes that can be used in neuromorphic applications such as biosensors \cite{liu2015k, wu2014k, wu2014l, wu2016a, zhou2014}. 

There is a very large amount of fascinating work being done in the materials science community to develop devices for neuromorphic systems out of novel materials in order to build smaller, faster, and more efficient neuromorphic devices.  Different materials for even a single device implementation can have wildly different characteristics.  These differences will propagate effects through the rest of the community, up through the device, high-level hardware, supporting software, model and algorithms levels of neuromorphic systems.  Thus, as a community, it is important that we understand what implications different materials may have on functionality, which will almost certainly require close collaborations with the materials science community moving forward. 

\subsection{Summary and Discussion}

In this section, we have looked at hardware implementations at the full device level, at the device component level, and at the materials level.  There is a significant body of work in each of these areas.  At the system level, there are fully functional neuromorphic systems, including both programmable architectures such as FPGAs and FPAAs, as well as custom chip implementations that are digital, analog, or mixed analog/digital.  A wide variety of novel device components beyond the basic circuit elements used in most device development have been utilized in neuromorphic systems.  The most popular new component that is utilized is the memristor, but other device components are becoming popular, including other memory technologies such as CBRAM and phase change memory, as well as spin-based components, optical components, and floating gate transistors.  There are also a large variety of materials being used to develop device components, and the properties of these materials will have fundamental effects on the way future neuromorphic systems will operate.  

\section{Supporting Systems}

In order for neuromorphic systems to be feasible as a complementary architecture for future computing, we must consider the supporting tools required for usability.  Two of the key supporting systems for neuromorphic devices are communication frameworks and supporting software.  In this section, we briefly discuss some of the work in these two areas.

\subsection{Communication}

\begin{figure*}[!t]
\centering
\subfloat[High-Level View]{\includegraphics[width=0.3\textwidth]{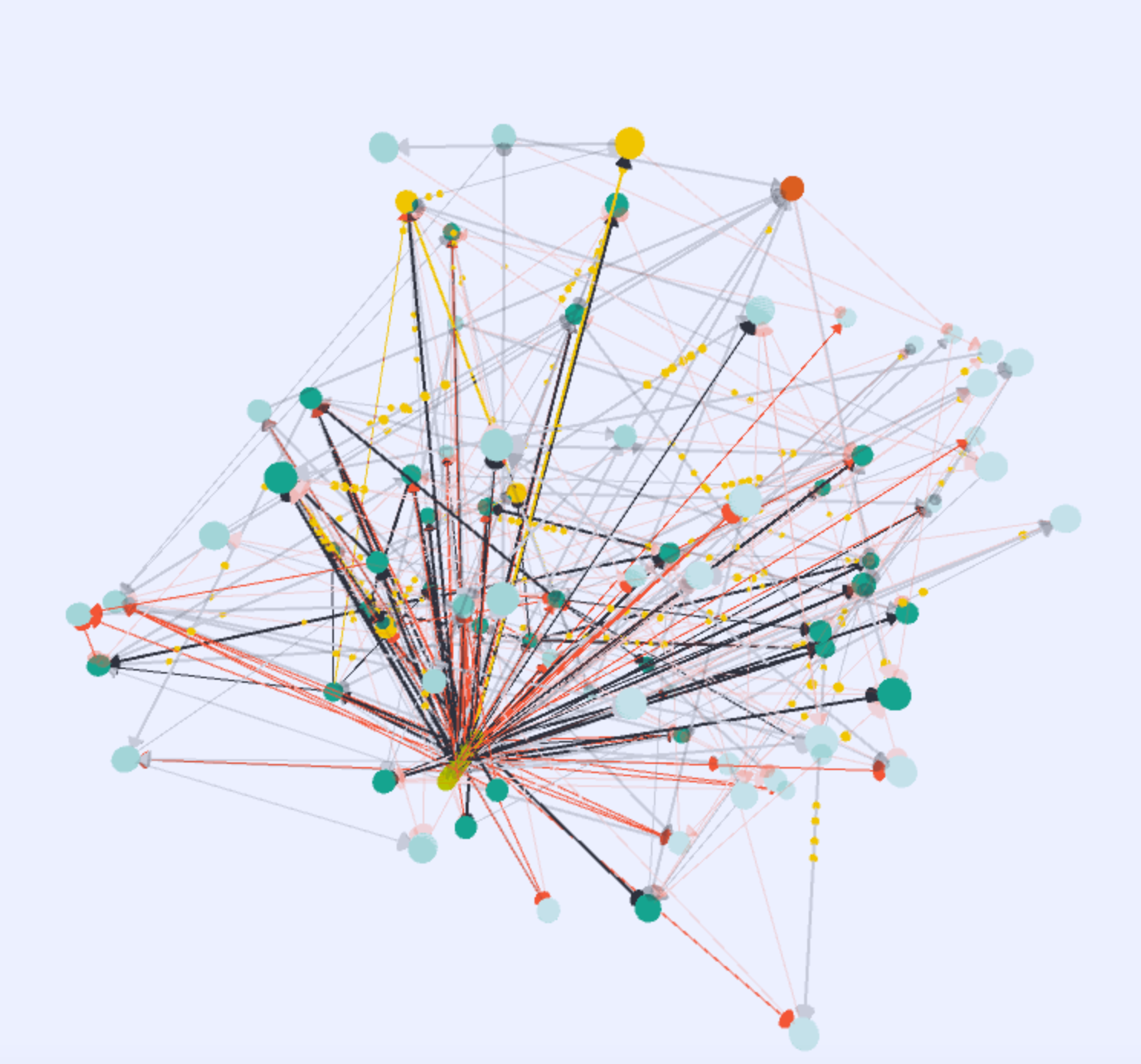}%
\label{fig:nidaviz}}
\hfil
\subfloat[Low-Level View]{\includegraphics[width=0.3\textwidth]{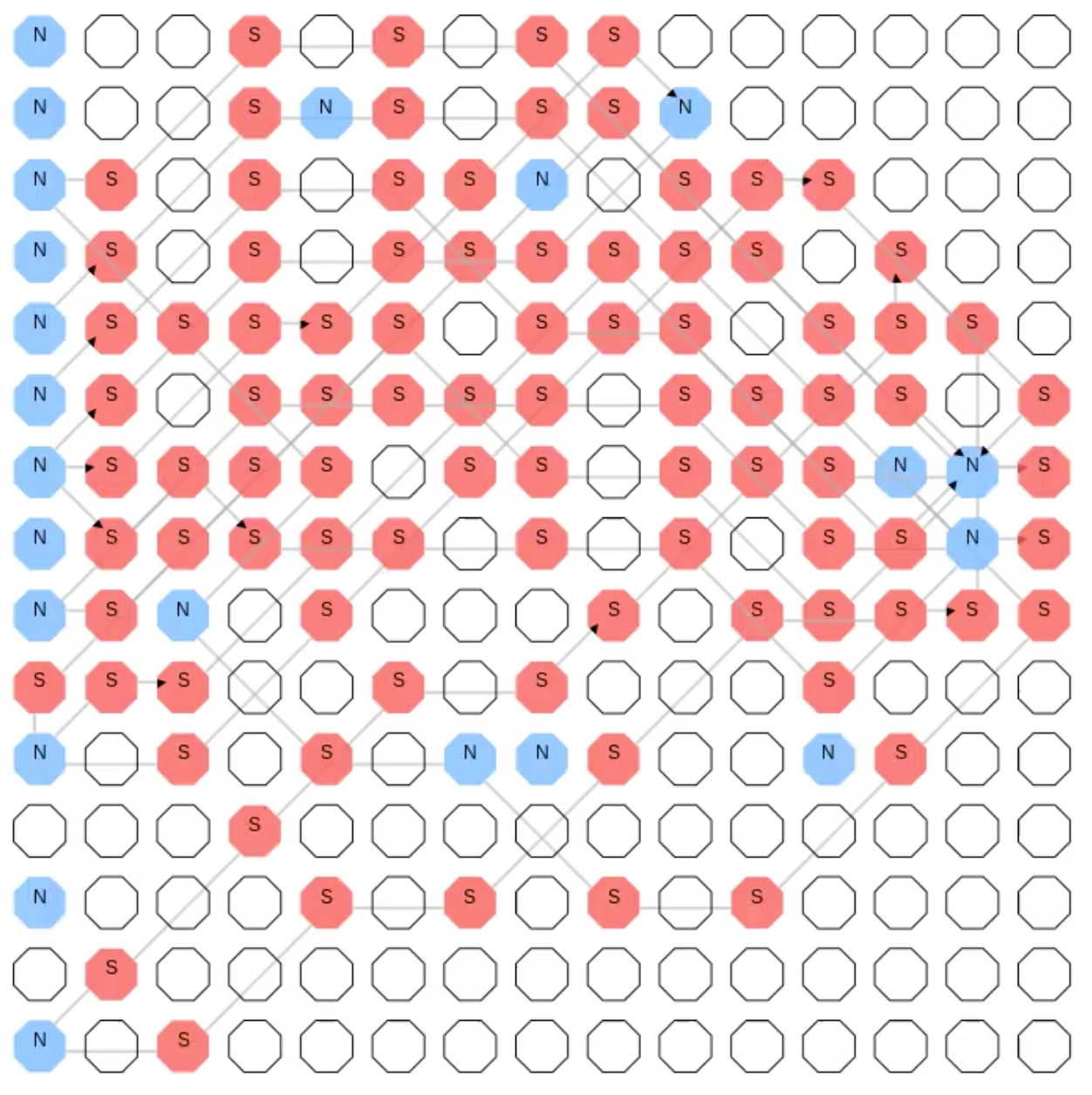}%
\label{fig:dannaviz}}
\caption{Example neuromorphic visualization tools, giving a high-level view of a spiking neural network model \cite{drouhard2014} and a low-level view of a network layout on a particular neuromorphic implementation \cite{disney2016}.}
\label{fig:viz_examples}
\end{figure*}

Communication for neuromorphic systems includes both intra-chip and inter-chip communication.  Perhaps the most common implementation of inter-chip communication is address event representation (AER) \cite{boahen1998, boahen1999, boahen2000, jablonski2015, liu2015i, merolla2007, merolla2014a, ramakrishnan2013, zamarreno2013}. In AER communication, each neuron has a unique address, and when a spike is generated that will traverse between chips, the address specifies to which chip it will go.  Custom PCI boards for AER have been implemented to optimize performance \cite{chicca2007, paz2006}. Occasionally, inter-chip communication interfaces will have their own hardware implementations, usually in the form of FPGAs \cite{thanasoulis2012, thanasoulis2012a, thanasoulis2014, thanasoulis2014a, scholze2012}.  SpiNNaker's interconnect system is one of its most innovative components; the SpiNNaker chips are interconnected in a toroidal mesh, and an AER communication strategy for inter-chip communication is used \cite{davies2012a, furber2006, navaridas2009, navaridas2012, navaridas2015, patterson2012, plana2008, plana2011, rast2008, rast2010, rast2012}.   It is the communication framework for SpiNNaker that enables scalability, allowing tens of thousands of chips to be utilized together to simulate activity in a single network.  A hierarchical version of AER utilizing a tree structure has also been implemented \cite{park2012a, park2016b}.  One of the key components of AER communication is that it is asynchronous.  In contrast, BrainScaleS utilizes an isynchronous inter-chip communication network, which means that events occur regularly \cite{philipp2007, philipp2009}. 

AER communication has also been utilized for on-chip communication \cite{deiss1999, sargeni2009}, but it has been noted that there are limits of AER for on-chip communication \cite{cassidy2011a}.  As such, there are several other approaches that have been used to optimize intra-chip communication.  For example, in early work with neuromorphic systems, buses were utilized for some on-chip communication systems \cite{mortara1994, mortara1995}.  In a later work, one on-chip communication optimization removed buses as part of the communication framework to improve performance \cite{brunelli2005}.  Vainbrand and Ginosaur examined different network-on-chip architectures for neural networks, including mesh, shared bus, tree, and point-to-point, and found network-on-chip multicast to give the highest performance \cite{vainbrand2010, vainbrand2011}. Ring-based communication for on-chip communication has also been utilized successfully \cite{pande2013, pande2013a}. Communication systems specifically for feed-forward networks have also been studied \cite{suzuki1989, hasan2013, hasan2013a}. 


One of the common beyond Moore's law era technologies to improve performance in communication that is being utilized across a variety of computing platforms (including traditional von Neumann computer systems) is three-dimensional (3D) integration.  3D integration has been utilized in neuromorphic systems even from the early days of neuromorphic, especially for pattern recognition and object recognition tasks \cite{duong1994, bermak2003}.  In more recent applications, 3D integration has been used in a similar way as it would be for von Neumann architectures, where memory is stacked with processing \cite{kim2016b}. It has also been utilized to stack neuromorphic chips.   Through silicon vias (TSVs) are commonly used to physically implement 3D integration approaches for neuromorphic systems \cite{clermidy2014, ehsan2015, ehsan2016, ehsan2016a, kim2016b}, 
partially because utilizing TSVs in neuromorphic systems help mitigate some of the issues that arise with using TSVs, such as parasitic capacitance \cite{joubert2012a}; however, other technologies have also been utilized in 3D integration, such as microbumps \cite{belhadj2014}.   3D integration is commonly used in neuromorphic systems with a variety of other technologies, such as memristors \cite{li2016g, wang2014k, wang2016m, manem2015, piccolboni2015}, phase change memory \cite{eryilmaz2013}, and CMOS-molecular (CMOL) systems \cite{ryan2009}. 






\subsection{Supporting Software}
\label{sec:supporting}

Supporting software will be a vital component in order for neuromorphic systems to be truly successful and accepted both within and outside the computing community.  However, there has not been much focus on developing the appropriate tools for these systems.  In this section, we discuss some efforts in developing supporting software systems for different neuromorphic implementations and use-cases.  

One important set of software tools consist of custom hardware synthesis tools \cite{achyuthan1994, achyuthan1994a, baptista2015, bayraktarouglu1999, braga2005, chen1990, mostafa2013, nigri1993, stewart2014}.  These synthesis tools typically take a relatively high level description and convert it to very low level representations of neural circuitry that can be used to implement neuromorphic systems.  They tend to generate application specific circuits.  That is, these tools are meant to work within the confines of a particular neuromorphic system, but also generate neuromorphic systems for particular applications.

A second set of software tools for neuromorphic systems are tools that are meant for programming existing neuromorphic systems.  These fall into two primary categories: mapping and programming.  Mapping tools are usually meant to take an existing neural network model representation, probably trained offline using existing methods such as back-propagation, and convert or map that neural network model to a particular neuromorphic architecture \cite{bruderle2011, dong2009, ehrlich2010, galluppi2012, galluppi2012a, gao2012, ji2016a, ji2016c, mayr2007, nigri1991, pino2010, sharp2011, sharp2011a, urgese2016, zhang2016c}.  These tools typically take into account restrictions associated with the hardware, such as connectivity restrictions or parameter value bounds, and make appropriate adaptations to the network representation to work within those restrictions.  

Programming tools, in contrast to mapping tools, are built so that a user can explicitly program a particular neuromorphic architecture \cite{brown2015, bruderle2007, disney2016, kasabov2015, kulkarni2012, neopane2016, scott2013, sheik2011, shinde2015, temam2011, tisan2015, wen2015a, wilby2015, wu2015e}. These can allow the user to program at a low level by setting different parameter and topology configurations, or by utilizing custom training methods built specifically for a particular neuromorphic architecture.  The Corelet paradigm used in TrueNorth programming fits into this category \cite{amir2013}.  Corelets are pre-programmed modules that accomplish different tasks.  Corelets can be used as building blocks to program networks for TrueNorth that solve more complex tasks.  There have also been some programming languages for neuromorphic systems such as PyNN \cite{davison2008, davison2010}, PyNCS \cite{stefanini2014}, and even a neuromorphic instruction set architecture \cite{hashmi2011}.  These languages have been developed to allow users to describe and program neuromorphic systems at a high-level. 

Software simulators have also been key in developing usable neuromorphic systems \cite{ade1992, atlas1989, bichler2013, disney2016, galluppi2010, ji2016, ji2016a, khalil2013, kolasa2015, kolasa2015a, kulkarni2012, plagge2016, plesser2007, preissl2012, rast2010, tisan2015, xia2016}.  Software-based simulators are vital for verifying hardware performance, testing new potential hardware changes, and for development and use of training algorithms. If the hardware has not been widely deployed or distributed, software simulators can be key to developing a user base, even if the hardware has not been fabricated beyond simple prototypes.  Visualization tools that show what is happening in neuromorphic systems can also be key to allowing users to understand how neuromorphic systems solve problems and to inspire further development within the field \cite{disney2016, dominguez2016, drouhard2014, kasabov2015, scott2013}.  These visualization tools are often used in combination with software simulations, and they can provide detailed information about what might be occurring at a low-level in the hardware.   Figure \ref{fig:viz_examples} provides two examples of visualizations for neuromorphic systems.  

\subsection{Summary}

When building a neuromorphic system, it is extremely important to think about how the neuromorphic system will actually be used in real computing systems and with real users.  Supporting systems, including communication on-chip and between chips and supporting software, will be necessary to enable real utilization of neuromorphic systems.  Compared to the number of hardware implementations of neuromorphic systems there are very few works that focus on the development of supporting software that will enable ease-of-use for these systems.  There is significantly more work to be done, especially on the supporting software side, within the field of neuromorphic computing.  It is absolutely necessary that the community develop software tools alongside hardware moving forward.  

\section{Applications}
\label{sec:applications}

\begin{figure}[!t]
\centering
\includegraphics[width=0.5\textwidth]{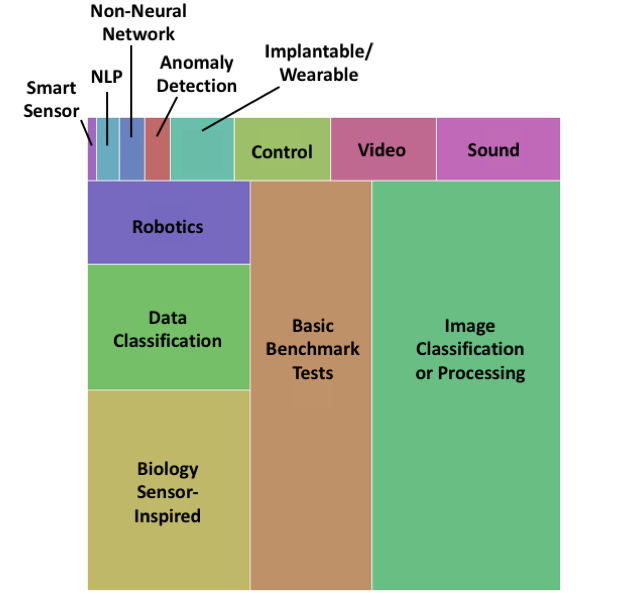}
\caption{Breakdown of applications to which neuromorphic systems have been applied.  The size of the boxes corresponds to the number of works in which a neuromorphic system was developed for that application.}
\label{fig:application_breakdown}
\end{figure}

The ``killer'' applications for neuromorphic computing, or the applications that best showcase the capabilities of neuromorphic computers and devices, have yet to be determined.  Obviously, various neural network types have been applied to a wide variety of applications, including image \cite{krizhevsky2012imagenet}, speech \cite{graves2013speech}, and data classification \cite{prechelt1994proben1}, control \cite{hagan2002introduction}, and anomaly detection \cite{ryan1998intrusion}.  Implementing neural networks for these types of applications directly in hardware can potentially produce lower power, faster computation, and a smaller footprint than can be delivered on a von Neumann architecture.  However, many of the application spaces that we will discuss in this section do not actually require any of those characteristics.  In addition, spiking neural networks have not been studied to their full potential in the way that artificial neural networks have been, and it may be that physical neuromorphic hardware is  required in order to determine what the killer applications for spiking neuromorphic systems will be moving forward.  This goes hand-in-hand with the appropriate algorithms being developed for neuromorphic systems, as discussed in Section \ref{sec:algorithms}.  Here, we discuss a variety of applications of neuromorphic systems.  We omit one of the major application areas, which is utilizing neuromorphic systems in order to study neuroscience via faster, more efficient simulation than is possible on traditional computing platforms.  Instead, we focus on other real-world applications to which neuromorphic systems have been applied.  The goal of this section is to provide a scope of the types of problems neuromorphic systems have successfully tackled, and to provide inspiration to the reader to apply neuromorphic systems to their own set of applications.  Figure \ref{fig:application_breakdown} gives an overview of the types of applications of neuromorphic systems and how popular they have been.

There are a broad set of neuromorphic systems that have been developed entirely based on particular sensory systems, and applied to those particular application areas.  The most popular of these ``sensing" style implementations are vision-based systems, which often have architectures that are entirely based on replicating various characteristics of biological visual systems \cite{adams2014, al2011, al2012, andreou1990, andreou1990a, andreou1994, andreou1995, andreou2016, arreguit1994, barranco2016, bartolozzi2004, bartolozzi2011, bartolozzi2011a, barzegarjalali2016, bevcanovic2005, boahen2005, chakradhar2010, chiang2004, chicca2006, chicca2007, choi2005, chotard2016, chua2014, conti2015, cosp2006, costas2007, culurciello2001, culurciello2003, dante2001, de2016a, debole2011, deiss1999, delbruck2007, douglas1994, douglas1995, eibensteiner2012, esser2013, etienne2006, fang1990, farabet2010, firouzi2015, firouzi2016, galluppi2012b, gao2016a, gupta2014, hammerstrom2003, han2007, han2012, han2013, han2014, han2015, han2015a, han2015b, han2016a, horiuchi1999, huyck2015, indiveri1994, indiveri1995, indiveri1999, indiveri1999a, indiveri2000, indiveri2000a, indiveri2001, indiveri2001b, indiveri2002b, indiveri2003a, indiveri2008, kang2015a, kavehei2014, kawasetsu2014, kestur2012, kramer1998, kudo2016, lichtsteiner2008, liu2001, liu2009, liu2010, luo2005, maher1989, mahowald1992, mandloi2014, markan2013, martel2016, martel2016a, meng2008, merolla2003, mishra2016, morris1998a, mueller2015, mueller2015a, netter2002, okuno2014, osswald2017, ozalevli2005, prieto2009, sandamirskaya2013, serrano2006, serrano2008, serrano2009, serrano2014, sharma2016, shi2003a, shimonomura2008, singh2016, snider2011, sonnleithner2011, suri2012, suri2013, tan2015, vogelstein2004a, vogelstein2007a, yagi2016, yang2006, yu2012a, zaghloul2004, zaghloul2004a, zamarreno2011}.  Though vision-based systems are by far the most popular sensory-based systems, there are also neuromorphic auditory systems \cite{chen2007, dominguez2016, dominguez2016a, douglas1995, glover1998, jimenez2016, jones2000, koickal2011, kuo1993, lazzaro1993, park2013a, piccolboni2015, saeki2002, sarpeshkar2006, sheik2012a, sheu1995, smith1998, smith2015, suri2012, suri2013, thakur2015, van1996, vanschaik2004, voutsas2007, zahn1996}, olfactory systems \cite{hsieh2012, hsieh2012a, imam2012a, koickal2006, koickal2007a, muir2005, pan2012, pearce2005, principe2001, shen2003, shen2004, tisan2010, tisan2015}, and somatosensory or touch-based systems \cite{pearson2006, rongala2015, ros2015, joshi2013, lee2015b, lee2017, rios2015}.   

Another class of applications for neuromorphic systems are those that either interface directly with biological systems or are implanted or worn as part of other objects that are usually used for medical treatment or monitoring \cite{harrer2016}.  A key feature of all of these applications is that they require devices that can be made very small and require very low-power.  Neuromorphic systems have become more popular in recent years in brain-machine interfaces \cite{boi2016, corradi2014b, corradi2015a, george2015, gordon2006, harrer2016, hsiao2006, jung2001, micera2015, nurse2016, thibeault2014, zumsteg2005}.  By their very nature, spike-based neuromorphic models communicate using the same type of communication as biological systems, so they are a natural choice for brain-machine or brain-computer interfaces.  Other wearable or implantable neuromorphic systems have been developed for pacemakers or defibrillator systems \cite{coggins1994, coggins1995, coggins1995a, leong1992, sun2011}, retina implants \cite{gaspar2015}, wearable fall detectors for elderly users \cite{vskoda2011}, and prosthetics \cite{zhang2011}.

Robotics applications are also very common for neuromorphic systems.  Very small and power efficient systems are often required for autonomous robots.  Many of the requirements for robotics, including motor control, are applications that have been successfully demonstrated in neural networks.  Some of the common applications of neuromorphic systems for robotics include learning a particular behavior \cite{daly2011, yamazaki2013}, locomotion control or control of particular joints to achieve a certain motion \cite{adhikari2015, arena2006, davies2010, de2002, eberhardt1992, gallagher2000, gallagher2001, iwata2016, lu2011, lu2013, lu2014, menon2014, niu2017, perez2014}, social learning \cite{belpaeme2016, cingolani2015}, and target or wall following \cite{dungen2005, furber2014, galluppi2014a, jackson1994, bellis2004}.  Thus far, in terms of robotics, the most common use of neuromorphic implementations is for autonomous navigation tasks \cite{acconcia2014, ames2012, arthur2012, azhar2002, beyeler2015, brownlow1991, buhlmeier1996, cawley2010, denk2013, howard2011, howard2012, howard2014, howard2014a, howard2015, hu2014, indiveri1997, koziol2014, pande2013a, reichel2005, reichel2005a, richert1991, rocke2007, roggen2003, stewart2016, tarassenko1991, tombs1993}.  In the same application space as robotics is the generation of motion through central pattern generators (CPGs).  CPGs generate oscillatory motions, such as those used to generate swimming motions in lampreys or walking gaits.  There are a variety of implementations of CPGs in neuromorphic systems \cite{ambroise2015, barron2013, basu2010a, donati2014, donati2016, gallagher2005, joucla2016, lee2007, lewis2000, lewis2003, orchard2008, patel2000a, ryckebusch1989, still2006, tenore2003, tenore2004, tenore2005, tenore2006, vogelstein2008, yang2012a}. 

Control tasks have been popular for neuromorphic systems because they typically require real-time performance, are often deployed in real systems that require small volume and low power, and have a temporal processing component, so they benefit from models that utilize recurrent connections or delays on synapses.  A large variety of different control applications have utilized neuromorphic systems \cite{dong2014, gallagher2008, hasanien2011, orlowska2011, tatikonda2008, soares2006, liu1999, liu1999a, bastos2006, sinha2011, sangeetha2013, bayindir2010, fukuda1992, jung2007, kim2004, kim2008, damle1997}, but by far the most common control test case is the cart-pole problem or the inverted pendulum task \cite{cawley2011, dean2016a, eguchi1991, grossi1995, jung2007, kim2004, pasero2004, rocke2008, schuman2016a, sitte2007}.  Neuromorphic systems have also been applied to video games, such as Pong \cite{arthur2012}, PACMAN \cite{galluppi2012a}, and Flappy Bird \cite{schuman2016a}. 

An extremely common use of both neural networks and neuromorphic implementations has been on various image-based applications, including edge detection \cite{glackin2009a, goldberg2001a, graf1990, hsu2014, kim2015g, li2015g, merkel2013, qi2014, roy2014b, salerno1999, sharad2012b,  sharad2012c, yakopcic2015a}, image compression \cite{bohrn2013, chasta2012, corinto2014, fang1992, pinjare2009, pinjare2012}, image filtering \cite{goldberg2001, harrer1992, ker1997, kim2012b, kim2014, krid2009, lee1990, lee1990a, lee1990b, martinez2003, querlioz2017, roychowdhury1996, shi2006}, image segmentation \cite{bernhard2006, caron2013, cosp1999, cosp2003, crebbin2005, graf1993, heittmann2002a, matolin2004, matolin2004a, nuno2012, perez1996, secco2015, seguin2015, vega2006, xiong2010}, and feature extraction \cite{ahmed2016a, andreou2016a, bichler2013, chen2007, chen2010a, fieres2004, georgiou2006, holleman2015, holleman2015a, indiveri2013a, kaulmann2005, knag2015, liu2006, lu2015, markan2007, sanchez2013, serrano2013, sharad2013}.  Image classification, detection, or recognition is an extremely popular application for neural networks and neuromorphic systems.    The MNIST data set, subsets of the data set, and variations of the data set has been used to evaluate many neuromorphic implementations \cite{afshar2015, ago2013, ahn2013a, ahn2014, al2015, al2015a, ambrogio2016a, ambrogio2016b, arthur2012, belhadj2014, bennett2016, bill2014, burger2014, burr2014, burr2015, burr2015a, chen2015b, chen2015c, chen2016b, cheng2016, chi2016, cloutier1994, cohen2016, das2015, diamond2015, disney2016, du2015b, esser2013, esser2015, fieres2006, fieres2006a, gamrat2015, gao2015, garbin2014, garbin2015, garbin2015a, gi2015, graf1993, han2016b, hasan2015, hashmi2011, hu2016, ji2016, ji2016a, ji2016b, kang2015, kataeva2015, kim2015c, kim2015e, kung2015, li2014, li2014d, liu2014, liu2015, liu2016a, liu2016c, merkel2015a, merolla2011, moon2016a, mrazek2016, naous2016, narasimman2016, neftci2013a, neftci2016, neftci2016a, negrov2016, neil2014, neil2016, neopane2016, park2016a, pedroni2013, pedroni2016, pietras2015, potok2016, querlioz2011, querlioz2012, querlioz2013, ramasubramanian2014, rothganger2015, sanni2015, sarwar2016, sengupta2015, sengupta2015a, sengupta2016e, sengupta2016f, sengupta2016j, shahsavari2016, sheik2016, shelby2015, sheri2015, shi2014a, sidler2016, srinivasan2016a, srinivasan2016b, starzyk2013, starzyk2014, strigl2010, stromatias2015, stromatias2015a, suri2015a, taha2009, tang2014, tissera2016, tsai2015, venkataramani2014, vianello2017, wang2015i, wang2015m, wen2016, woo2016a, woods2014, woods2015, yakopcic2014a, yakopcic2015a, yakopcic2016, yang2016c, yepes2016, zamanidoost2015, zhang2015e, zhang2016b, zhang2016c}.  Other digit recognition tasks \cite{achyuthan1994, achyuthan1994a, alippi1991a, boser1991, chen2014b, chu2015, darwish2016, duan2015, duan2016a, lorenzi2015, masmoudi1999, merkel2015, mitra2009, naous2016a, rast2012a, rothganger2015, sackinger1991, sackinger1992, sato1993, tam1992, tomberg1990, wu2015, yan2016, yang2016c, zhang2017, zhao2012, zhu2016} and general character recognition tasks \cite{agranat1990, basaglia1995, bishop2010, covi2016a, diep2014, domingos2005, ehkan2014, foruzandeh1999a, han2010, han2010a, hoffman2006, horita2015, hu2012, hu2012a, hu2014a, jackel1990, jackel1990a, jang2016, jang2016a, khan2006, kim1992, kim1993, kim1999, kim2012a, kim2013a, kim2014a, kim2015a, maeda2003, mansour2011, morns1999, nirmaladevi2015, oh2004, pino2010, pino2012, qiu2013, qiu2014, qiu2015, reyneri1991, rice2009, sarwar2016, sengupta2016d, sharad2012, sharad2012a, sharad2012c, sheri2014, soleimani2012, soltiz2013, sousa2014, taha2010, tam1990, tang1997, tarkov2015a, tarkov2016, tarkov2016a, truong2014a, wang1991, wang2014s, wang2016g, white1992, yu2006, zamarreno2013, el2003, graf1987, graf1988, graf1995, kim1998, lucas1991, yamakawa1996} have also been very popular.  Recognition of other patterns such as simple shapes or pixel patterns have also been used as applications for neuromorphic systems \cite{cilingiroglu1990, covi2016, damak2006, djahanshahi1996a, eryilmaz2013, giulioni2008a, giulioni2009, giulioni2015, hikawa1999, hikawa2001, hikawa2002, hikawa2003, hikawa2003a, huayaney2011, iakymchuk2012, ielmini2016, indiveri2015c, kanazawa2003, kaneko2013, kaneko2014, kim1995, kim1995a, kim2015d, kinget1995, kuzum2011a, kuzum2012, leiner2008, li2012, linares1992, liu2012, liu2013, liu2013b, liyanagedera2016, lont1992, maundy1991a, mitra2007, moreno2009, nishitani2014, payvand2014, plaskonos1993, prezioso2015, prezioso2016, prezioso2016d, ranjan2016, rosado2011, schmid1999, serrano1996, taha2014a, tanaka2009, tarkov2015, tuazon1993, upegui2003, wang2014r, wang2016m, wen2014, wozniak2016, xu2015a, yakopcic2013, yao2015, yogendra2015, zamanlooy2012}.  

\begin{figure}[!t]
\centering
\includegraphics[width=0.5\textwidth]{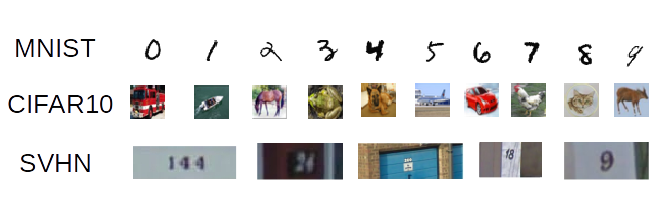}
\caption{Examples from different image data sets (MNIST \cite{lecun1998mnist}, CIFAR10 \cite{krizhevsky2014cifar}, and SVHN \cite{netzer2011svhn}) to which neuromorphic systems have been applied for classification purposes.}
\label{fig:image_dataset_examples}
\end{figure}

Other image classification tasks that have been demonstrated on neuromorphic systems include classifying real world images such as traffic signs \cite{fieres2006, garbin2015a, vitabile2005}, face recognition or detection \cite{adhikari2015, chevitarese2016, farabet2009, ibrayev2016, knag2016, mahdiani2012, ramasubramanian2014, sarwar2016, venkataramani2014}, car recognition or detection \cite{adhikari2012}, detecting air pollution in images \cite{bolouri1995}, detection of manufacturing defects or defaults \cite{onorato1994}, hand gesture recognition \cite{aibe2004, sofatzis2014}, human recognition \cite{perez2010}, object texture analysis \cite{andreopoulos2016}, and other real world image recognition tasks \cite{aibe2002, erkmen2013, kim2016b, roska2012}.  There are several common real-world image data sets that have been evaluated on neuromorphic systems, including the CalTech-101 data set \cite{farabet2011, merkel2014c}, the Google Street-View House Number (SVHN) data set \cite{mrazek2016, sarwar2016, tissera2016, vianello2017, venkataramani2014}, the CIFAR10 data set \cite{cao2015, chung2016, esser2016, li2016b, ramasubramanian2014, tissera2016, venkataramani2014}, and ImageNet \cite{chi2016}.  Evaluations of how well neuromorphic systems implement AlexNet, a popular convolutional neural network architecture for image classification, have also been conducted \cite{chen2016c, suda2016}.  Examples of images from the MNIST data set, the CIFAR10, and the SVHN data set are given in Figure \ref{fig:image_dataset_examples} to demonstrate the variety of images that neuromorphic systems have been used to successfully classify or recognize.  

A variety of sound-based recognition tasks have also been considered with neuromorphic systems.  Speech recognition, for example, has been a common application for neuromorphic systems \cite{andreou2016, bayraktarouglu1999, bor1996, botros1994, canas2006, cassidy2007, churcher1993, coath2013, cornu1994, ferrer2004, gatt2000, gomez2011, hammerstrom1990, hammerstrom1991, modi2005, ortigosa2003a, ortigosa2006, ortigosa2006a, rafique2016, schrauwen2008, such2015, suri2015, trujillo2012, truong2014, truong2016, tsai2016, van1994, vandoorne2014, wang2016l}.  Neuromorphic systems have also been applied to music recognition tasks \cite{amir2013, cerezuela2015, esser2013}.  Both speech and music recognition tasks may require the ability to process temporal components, and may have real-time constraints.  As such, neuromorphic systems that are based on recurrent or spiking neural networks that have an inherent temporal processing component are natural fits for these applications.  Neuromorphic systems have also been used for other sound-based tasks, including speaker recognition \cite{esser2013}, distinguishing voice activity from noise \cite{esser2016}, and analyzing sound for identification purposes \cite{glover1999, glover2002}.  Neuromorphic systems have also been applied to noise filtering applications as well to translate noisy speech or other signals into clean versions of the signal \cite{almeida1996, cho1999, choi1993a, choi1993b}. 

Applications that utilize video have also been common uses of neuromorphic systems.  The most common example for video is object recognition within video frames \cite{akolkar2015, akopyan2015, cassidy2014, chen2011, garbin2013, kasabov2015, khosla2013, khosla2014, kim2010, madokoro2013, merolla2014, orchard2015, rodriguez2015, shin2016, suri2012b, suri2012c, suri2013, suri2013a, suri2013b}.  This application does not necessarily require a temporal component, as it can analyze video frames as images.  However, some of the other applications in video do require a temporal component, including motion detection \cite{brosch2016, giulioni2016, higgins1999, higgins2000, wang2010}, motion estimation \cite{garcia2014, torralba1999, van1994}, motion tracking \cite{reverter2015a, krips2002, liu2015g, oniga2007}, and activity recognition \cite{kung2016, oniga2004}. 

Neuromorphic systems have also been applied to natural language processing (NLP) tasks, many of which require recurrent networks. Example applications in NLP that have been demonstrated using neuromorphic systems include sentence construction \cite{ahmed2016a}, sentence completion \cite{li2014c, li2015h, pino2010}, question subject classification  \cite{diehl2016}, sentiment analysis \cite{diehl2016a}, and document similarity \cite{marchese2015}.  

Tasks that require real-time performance, ability to deploy into an environment with a small footprint, and/or low power are common use cases for neuromorphic systems.  Smart sensors are one area of interest, including humidty sensors \cite{islam2006}, light intensity sensors \cite{patra2006}, and sensors on mobile devices that can be used to classify and authenticate users \cite{phillips2016}.  Similarly, anomaly detectors are also applications for neuromorphic systems, including detecting anomalies in traffic \cite{chen2013b}, biological data \cite{otahal2016}, and industrial data \cite{otahal2016},  applications in cyber security \cite{pino2014, shi2015b}, and fault detection in diesel engines \cite{shreejith2016} and analog chips \cite{mosin2015, mosin2016}. 



General data classification using neuromorphic systems has also been popular.  There are a variety of different and diverse application areas in this space to which neuromorphic systems have been applied, including accident diagnosis \cite{syiam2003}, cereal grain identification \cite{dolenko1993, dolenko1995}, computer user analysis \cite{draghici1998, draghici1998a}, driver drowsiness detection \cite{khalil2013}, gas recognition or detection \cite{al2001, alizadeh2008, benrekia2009}, product classification \cite{faiedh2004}, hyperspectral data classification \cite{christiani2016}, stock price prediction \cite{yasunaga1991}, wind velocity estimation \cite{zhang2005}, solder joint classification \cite{halgamuge1994}, solar radiation detection \cite{shaari2008}, climate prediction \cite{acosta2001}, and applications within high energy physics \cite{denby2003, vitabile2005}.  Applications within the medical domain have also been popular, including coronary disease diagnosis \cite{izeboudjen2007}, pulmonary disease diagnosis \cite{economou1994}, deep brain sensor monitoring \cite{zhang2015g}, DNA analysis \cite{stepanova2007}, heart arrhythmia detection \cite{khalil2013}, analysis of electrocardiogram (ECG) \cite{izeboudjen1999, ozdemir2011, qi2014, sun2012}, electroencephalogram (EEG) \cite{kudithipudi2015, park2015, polepalli2016, shen2016}, and electromyogram (EMG) \cite{bu2004, kudithipudi2015} results, and pharmacology applications \cite{pastur2016}.  A set of benchmarks from the UCI machine learning repository \cite{blake1998uci} and/or the Proben1 data set \cite{prechelt1994proben1, beuchat1998a, lotrivc2012, modi2006} have been popular in both neural network and neuromorphic systems, including the following data sets:  building \cite{liu2014, lotrivc2011}, connect4 \cite{liu2015, liu2016a}, gene \cite{liu2014, liu2015, liu2016a, lotrivc2011}, glass \cite{braendler2002a, chen2015c, girau2001}, heart \cite{hussain2015, lotrivc2011}, ionosphere \cite{chen2015c, decherchi2012, hussain2015}, iris \cite{bako2010, braendler2002, chen2012a, chen2015c, ferreira2010, ghani2012, girones2005, goknar2012, halgamuge1994, hsieh2013, johnston2005, low2006, merchant2008, merchant2010, nuno2009, pandya2005, rosenthal2016, santos2011, schuman2016, soudry2015}, lymphography \cite{liu2015, liu2016a}, mushroom \cite{liu2014, liu2015, liu2016a}, phoneme \cite{girones2005}, Pima Indian diabetes \cite{braendler2002, braendler2002a, girau2006a, lotrivc2011, schuman2016, suri2015b}, semeion \cite{ferreira2010}, thyroid \cite{liu2014, liu2015, liu2016a, lotrivc2011}, wine \cite{chen2012a, chen2015c, rosenthal2016, schuman2016}, and Wisconsin breast cancer \cite{bako2010, braendler2002a, cawley2011, decherchi2012, ghani2012, hasan2014, hsieh2013, hussain2015, liu2015, liu2016a, lotrivc2011, pandya2005, rosenthal2016, schuman2016, soudry2015}. These data sets are especially useful because they have been widely used in the literature and can serve as points of comparison across both neuromorphic systems and neuromorphic models. 

There are a set of relatively simple testing tasks that have been utilized in evaluating neuromorphic system behaviors, especially in the early development of new devices or architectures.  In the early years of development, the two spirals problem was a common benchmark \cite{beiu1995a, beiu1996, draghici1998, draghici1998a, petridis1995}.  $N$-bit parity tasks have also been commonly used \cite{adhikari2012, adhikari2014, adhikari2015, alspector1993, beiu1996, devi2010, dolenko1993, dolenko1993a, dolenko1995, duong1992, hikawa2003, hohmann2002, jayakumar1992, lu2002b, lu2003, merchant2010, nirmaladevi2015, pino2014, saichand2008, schemmel2001, schemmel2004a, valle1994}.  Basic function approximation has also been a common task for neuromorphic systems, where a mathematical function is specified and the neuromorphic system is trained to replicate its output behavior \cite{abdelbaki2000, al1998, alippi1991, bahoura2011a, basu2013, bayraktarouglu1999, blake1997, blake1998, braga2005, cauwenberghs1994, cauwenberghs1996, chiang2015, cho1998, corneil2012, corneil2012a, corradi2014, davies2013, del2008, eickhoff2006, farsa2015, farsa2015a, ferreira2010, ferrer2004, frye1992, glackin2005a, glackin2009, hu2014d, korkmaz2013, lin2008, lin2009, maeda2003, mundie1994, porrmann2002, wakamura2003}.  Temporal pattern recall or classification \cite{ang2011, arthur2006, azghadi2015, banerjee2015, basu2010a, bofill2003, chen2014a, esser2013, galluppi2014, hafliger1997, kanazawa2004, mahowald1997, stewart2014, tuma2016, zhao2015a} and spatiotemporal pattern classification \cite{barzegarjalali2016b, berdan2016, deng2015, etienne1994, hussain2016, kasabov2015, kasabov2016, lee2014b, mitra2007a, mumford1992, pantazi2016, polepalli2016a, rasche2007, roy2013, roy2014, scott2013, serrano2005, sheik2013, wang2009, wang2013, wang2014b} have also been popular with neuromorphic systems, because they demonstrate the temporal processing characteristics of networks with recurrent connections and/or delays on the synapses.  Finally, implementing various simple logic gates have been common tests for neural networks and neuromorphic systems, including AND \cite{aggarwal2012, bennett2015, gacem2013, gale2013, hikawa1995, hikawa2003a, kavehei2011, koosh2001, koosh2002, li2012a, soltiz2013, wang2012a, wolpert1992}, OR \cite{bennett2015, gacem2013, hikawa2003a, li2012a, makwana2013, soltiz2013, wang2012a, wang2016i}, NAND \cite{aunet2004, aunet2005, bennett2015, demin2015, demin2016, emelyanov2015, gacem2013, joye2007, liao2011, makwana2013, soltiz2013, wang2013d}, NOR \cite{aunet2004, aunet2005, bennett2015, demin2015, demin2016, emelyanov2015, gacem2013, joye2007, kavehei2011, liao2011, makwana2013, soltiz2013, wang2013d}, NOT \cite{gale2013, joye2007}, and XNOR \cite{kavehei2011, montalvo1997, sharifi2010, soltiz2013, wang2013d}.  XOR has been especially popular because the results are not linearly separable, which makes it a good test case for more complex neural network topologies \cite{abutalebi1998, achyuthan1994, achyuthan1994a, bayraktarouglu1999, berg1996, bo1999, braendler2002, card1994a, card1995, cawley2011, chabi2015, cho1996, choi1993, diotalevi2000, domingos2005, draghici1998, draghici1998a, dytckov2014, ebong2011, emelyanov2016, foruzandeh1999a, gadea2000, gale2017, ghani2011, girau2006, girones2005, hikawa2003a, hirotsu1993, ishii1992, izeboudjen2007, jabri1992, kavehei2011, khalil2013, kim2015f, kollmann1996, koosh2001, koosh2002, kosaka1995, krcma2015a, kudithipudi2014, lakshmi2013, li2012a, liu2005a, lu2000a, lu2001b, lysaght1994, madani2005, maeda1993, maeda1995, maeda1999, maeda2003, maher2006, maliuk2012, maliuk2014, maliuk2014a, mathia2002, maya2000, meador1989, meador1990, merchant2006, merkel2014b, mirhassani2003, mirhassani2004, mirhassani2005, mirhassani2007, montalvo1997, montalvo1997a, morgan2009, morie1994, morie1999, morie2000, moussa2006, murtagh1992, nambiar2014, nichols2002, nishitani2015, ota1999, pande2013a, pandya2005, pino2014, rani2007, salam1990, savran2003, sharifi2010, soelberg1994, soltiz2012, soltiz2013, tang1993, valle1994, vandoorne2014, wang1993, wang2012a, wang2013d, wolpert1992, yasunaga1993}.

\begin{figure*}
\begin{center}
\includegraphics[width=1.0\textwidth]{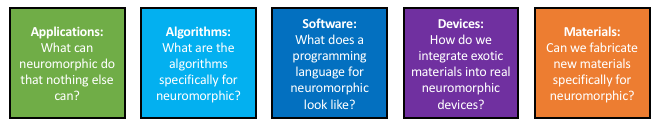}
\caption{Major neuromorphic computing research challenges in different fields.}
\label{fig:grand_challenges}
\end{center}
\end{figure*}

In all of the applications discussed above, neuromorphic architectures are utilized primarily as neural networks.  However, there are some works that propose utilizing neuromorphic systems for non-neural network applications.  Graph algorithms have been one common area of application, as most neuromorphic architectures can be represented as graphs.  Example graph algorithms include maximum cut \cite{bojnordi2016}, minimum graph coloring \cite{harmanani2010, roy1999}, traveling salesman \cite{sato2003, varma2002}, and shortest path \cite{aibara1991}. Neuromorphic systems have also been used to simulate motion in flocks of birds \cite{araujo2014} and for optimization problems \cite{rodriguez1990}. Moving forward, we expect to see many more use cases of neuromorphic architectures in non-neural network applications, as neuromorphic computers become more widely available and are considered simply as a new type of computer with specific characteristics that are radically different from the characteristics traditional von Neumann architecture.

\section{Discussion: Neuromorphic Computing Moving Forward}

In this work, we have discussed a variety of components of neuromorphic systems: models, algorithms, hardware in terms of full hardware systems, device-level components, new materials, supporting systems such as communication infrastructure and supporting software systems, and applications of neuromorphic systems.  There has clearly been a tremendous amount of work up until this point in the field of neuromorphic computing and neural networks in hardware.  Moving forward, there are several exciting research directions in a variety of fields that can help revolutionize how and why we will use neuromorphic computers in the future (Figure \ref{fig:grand_challenges}).

From the machine learning perspective, the most intriguing question is what the appropriate training and/or learning algorithms are for neuromorphic systems.  Neuromorphic computing systems provide a platform for exploring different training and learning mechanisms on an accelerated scale.  If utilized properly, we expect that neuromorphic computing devices could have a similar effect on increasing spiking neural network performance as GPUs had for deep learning networks.  In other words, when the algorithm developer is not reliant on a slow simulation of the network and/or training method, there is much faster turn around in developing effective methods.  However, in order for this innovation to take place, algorithm developers will need to be willing to look beyond traditional algorithms such as back-propagation and to think outside the von Neumann box.  This will not be an easy task, but if successful, it will potentially revolutionize the way we think about machine learning and the types of applications to which neuromorphic computers can be applied.

From the device level perspective, the use of new and emerging technologies and materials for neuromorphic devices is one of the most exciting components of current neuromorphic computing research.  With today's capabilities in fabrication of nanoscale materials, many novel device components are certain to be developed.  Neuromorphic computing researchers at all levels, including models and algorithms, should be collaborating with materials scientists as these new materials are developed in order to customize them for use in a variety of neuromorphic use cases.  Not only is there potential for extremely small, ultra-fast neuromorphic computers with new technologies, but we may be able to collaborate to build new composite materials that elicit behaviors that are specifically tailored for neuromorphic computation.

From the software engineering perspective, neuromorphic computers represent a new challenge in how to develop the supporting software systems that will be required for neuromorphic computers to be usable by non-experts in the future.  The neuromorphic computing community would greatly benefit from the inclusion of more software engineers as we continue to develop new neuromorphic devices moving forward, both to build supporting software for those systems, but also to inform the design themselves.  Once again, neuromorphic computers require a totally different way of thinking than traditional von Neumann architectures.  Building out programming languages specifically for neuromorphic devices wherein the device is not utilized as a neural network simulator but as a special type of computer with certain characteristics (e.g., massive parallelism and collocated memory and computation elements) is one way to begin to attract new users, and we believe such languages will be extremely beneficial moving forward.

From the end-user and applications perspective, there is much work that the neuromorphic community needs to do to develop and communicate use cases for neuromorphic systems.  Some of those use cases include as a neuromorphic co-processor in a future heterogeneous computer, as smart sensors or anomaly detectors in Internet of Things applications, as extremely low power and small footprint intelligent controllers in autonomous vehicles, as \textit{in situ} data analysis platforms on deployed systems such as satellites, and many other application areas.  The potential to utilize neuromorphic systems for real-time spatiotemporal data analysis or real-time control in a very efficient way needs to be communicated to the community at large, so that those that have these types of applications will think of neuromorphic computers as one solution to their computing needs.   

There are clearly many exciting areas of development for neuromorphic computing.  It is also clear that neuromorphic computers could play a major role in the future computing landscape if we continue to ramp up research at all levels of neuromorphic computing, from materials all the way up to algorithms and models.  Neuromorphic computing research would benefit from coordination across all levels, and as a community, we should encourage that coordination to drive innovation in the field moving forward.  


\section{Conclusion}

In this work, we have given an overview of past work in neuromorphic computing.  The motivations for building neuromorphic computers have changed over the years, but the need for a non-von Neumann architecture that is low-power, massively parallel, can perform in real time, and has the potential to train or learn in an on-line fashion is clear.  We discussed the variety of neuron, synapse and network models that have been used in neuromorphic and neural network hardware in the past, emphasizing the wide variety of selections that can be made in determining how the neuromorphic system will function at an abstract level.  It is not clear that this wide variety of models will ever be narrowed down to one all encompassing model in the future, as each model has its own strengths and weaknesses.  As such, the neuromorphic computing landscape will likely continue to encompass everything from feed-forward neural networks to highly-detailed biological neural network emulators.  

We discussed the variety of training and learning algorithms that have been implemented on and for neuromorphic systems.  Moving forward, we will need to address building training and learning algorithms specifically for neuromorphic systems, rather that adapting existing algorithms that were designed with an entirely different architecture in mind.  There is great potential for innovation in this particular area of neuromorphic computing, and we believe that it is one of the areas for which innovation will have the most impact.  We discussed high-level hardware views of neuromorphic systems, as well as the novel device-level components and materials that are being used to implement them.  There is also significant room for continued development in this area moving forward.  We briefly discussed some of the supporting systems for neuromorphic computers, such as supporting software, of which there is relatively little and from which the community would greatly benefit.  Finally, we discuss some of the applications to which neuromorphic computing systems have been successfully applied. 

The goal of this paper was to give the reader a full view of the types of research that has been done in neuromorphic computing across a variety of fields.  As such, we have included all of the references in this version. We hope that this work will inspire others to develop new and innovative systems to fill in the gaps with their own research in the field and to consider neuromorphic computers for their own applications.  


%

%

\section*{Acknowledgment}

This material is based upon work supported in part by the U.S. Department of Energy, Office of Science, Office of Advanced Scientific Computing Reserach, under contract number DE-AC05-00OR22725.   Research sponsored in part by the Laboratory Directed Research and Development Program of Oak Ridge National Laboratory, managed by UT-Battelle, LLC, for the U. S. Department of Energy.

\ifCLASSOPTIONcaptionsoff
  \newpage
\fi



\bibliographystyle{IEEEtran}
\end{document}